\documentclass[preprint,12pt]{elsarticle}

\usepackage{amsmath,amsfonts,amssymb,amsthm,mathtools,enumitem}
\usepackage[export]{adjustbox}
\usepackage{comment,hyperref,tikz,pifont,multirow}
\usepackage{algorithmic}
\usepackage{algorithm}
\usepackage{array,soul}
\usepackage[caption=false,font=normalsize,labelfont=sf,textfont=sf]{subfig}
\usepackage{textcomp}
\usepackage{stfloats,colortbl,xcolor,graphicx}
\usepackage{url}
\usepackage{verbatim}
\usepackage{rsfso}
\usepackage{enumerate}
\usepackage{subcaption}

\newtheorem{lemma}{Lemma}

\newtheorem{theorem}{Theorem}
\newtheorem{corollary}{Corollary}

\journal{Elsevier Signal Processing}
\usepackage{setspace} 

\usepackage[english]{babel}
\usepackage{bbm}
\usepackage[normalem]{ulem}

\newcommand{\xpcttn}{\mathbb{E}}
\newcommand{\xpcttnb}[1]{\ensuremath{\mathbb{E} \left[ {#1} \right]}}
\newcommand{\bigo}{\mathcal{O}}
\newcommand{\bigob}[1]{\ensuremath{\mathcal{O} \left( {#1} \right)}}
\newcommand{\mycomp}[2]{\mathrel{\overset{\makebox[0pt]{\mbox{\normalfont\footnotesize #1}}}{#2}}}
\newcommand\numberthis{\addtocounter{equation}{1}\tag{\theequation}}
\newcommand{\indctr}{\mathbbm{1}}

\definecolor{Gray}{gray}{0.93}

\newcommand{\customtitle}[1]{
    \begin{center}
        \Huge \textbf{#1}
    \end{center}
    \vspace{1cm} 
}

\begin{document}
\begin{frontmatter}

\title{Signal Reconstruction from Samples at Unknown Locations with Application to 2D Unknown View Tomography}

\author[inst2]{Sheel Shah}
\affiliation[inst2]{organization={Department of Electrical Engineering, Indian Institute of Technology, Bombay}}
\affiliation[inst3]{organization={Walmart Global Tech, Bangalore}}
\affiliation[inst1]{organization={Department of Computer Science and Engineering, Indian Institute of Technology, Bombay}}
\author[inst2]{Kaishva Shah}
\author[inst3]{Karthik S. Gurumoorthy}
\author[inst1]{Ajit Rajwade}
\ead{ajitvr@cse.iitb.ac.in}

\begin{abstract}
It is well known that a band-limited signal can be reconstructed from its uniformly spaced samples if the sampling rate is sufficiently high. More recently, it has been proved that one can reconstruct a 1D band-limited signal even if the exact sample locations are unknown, but given a uniform distribution of the sample locations and their ordering in 1D. In this work, we extend the analytical error bounds in such scenarios for quasi-bandlimited (QBL) signals, and for the case of arbitrary but known sampling distributions. We also prove that such reconstruction methods are resilient to a certain proportion of errors in the specification of the sample location ordering. We then express the problem of tomographic reconstruction of 2D images from 1D Radon projections under unknown angles (2D UVT) with known angle distribution, as a special case for reconstruction of QBL signals from samples at unknown locations with known distribution. Building upon our theoretical background, we present asymptotic bounds for 2D QBL image reconstruction from 1D Radon projections in the unknown angles setting, and present an extensive set of simulations to verify these bounds in varied parameter regimes. To the best of our knowledge, this is the first piece of work to perform such an analysis for 2D UVT and explicitly relate it to advances in sampling theory, even though the associated reconstruction algorithms have been known for a long time.
\end{abstract}
\end{frontmatter}


\section{Introduction}
The classical result of Shannon and Nyquist states that a bandlimited signal $g(t)$ can be accurately reconstructed from its equi-spaced time domain samples $\{g(iT)\}_{i=1}^N$, provided the sampling rate $1/T$ exceeds twice the maximum frequency of the signal (Nyquist rate). The signal reconstruction proceeds by the well-known Whittaker-Shannon interpolation formula given by $\hat{g}(t) = \sum_{i=1}^N g(iT) \textrm{sinc} \left(\frac{t-iT}{T}\right)$ where $i$ is the sample index. This result has later been extended to handle the case of non-uniformly spaced time domain samples. In both cases, the precise location of the samples is assumed to be known. However, this assumption is not valid in applications such as remote sensing, robotics, and image reconstruction. Hence, there has been previous research in signal and image reconstruction, in the case where the sampling locations are known imprecisely \cite{Nordio2008} (for example, as small random perturbations of points on a deterministic grid). They find their applications in Fourier Transform Spectrometry, and compressive reconstruction of MRI or diffusion MRI (DMRI) data \cite{Pandotra2019,Vaish2022} due to Fourier frequency mis-specification arising from hardware errors such as gradient delays in the MR machine. 

All the aforementioned examples assume that the specified sampling locations are only \emph{imprecise}. However, in some situations, the sampling locations may be \emph{completely unknown}. In such a setting, a signal reconstruction technique with analytical error bounds for band-limited signals with samples drawn from a uniform distribution and with known ordering, has been presented in  \cite{Kumar2015}. An image processing application which requires reconstruction from samples at unknown locations is unknown view tomography (UVT), which involves reconstruction of an image from its tomographic projections acquired under unknown viewing parameters \cite{Basu2000_a}. The primary aim of this paper is: (\textit{i}) to extend the theory from \cite{Kumar2015} to handle quasi bandlimited signals, arbitrary sampling distributions, and possibly unknown sample ordering; (\textit{ii}) to tie the UVT problem to such recent developments in sampling theory for signal reconstruction from samples at unknown locations, and (\textit{iii}) to develop, analyze and empirically verify theoretical performance bounds for image reconstruction in this application. To the best of our knowledge, this is the first analysis of UVT in the literature from the perspective (\textit{i}). 

\section{Related Work and Preliminaries}
\label{sec:prelim}
In this section, we briefly describe some existing techniques for bandlimited signal reconstruction from samples at unknown locations (USL), then present a brief overview of tomographic reconstruction, and finally tie together the UVT application with methods of signal reconstruction from USL.  

\subsection{Signal Reconstruction from Samples at Unknown Locations} 
There exist numerous applications in signal and image processing requiring reconstruction from USL. For instance, in remote sensing, the precise location of sensors being deployed to measure a certain spatial field maybe unknown due to sensor movement over time~\cite{Kumar2015}. In robotics, a robot may be attempting to localize itself in an environment whose precise map is unknown, which is called the simultaneous localization and mapping problem (SLAM)~\cite{Smith1986}. In cryo-electron microscopy \cite{Frank2006}, one needs to reconstruct an image (up to unknown global rotation) from its tomographic projections acquired at unknown angles. This is an important motivation for studying the UVT problem. In the problem of structure from motion in computer vision \cite{Trucco1998}, the aim is to reconstruct the 3D object's surface (up to unknown global rotation) using photographs acquired from unknown camera viewpoints. 

Theoretical analysis of signal or image reconstruction from USL poses significant challenges as the Whittaker-Shannon interpolation formula does not directly apply to this setting. The quantum of past works studying the problem of signal or image reconstruction from USL appears to be modest in size. In \cite{Marziliano2000}, numerical algorithms are proposed for the reconstruction of discrete-time bandlimited signals from USL, for sampling locations from sets with jitter around a uniform set, and non-uniform periodic sets of sampling locations. In \cite{Kumar2015}, it is shown that reconstruction of a bandlimited signal $g(t)$ from sample values $\{g(t_i)\}_{i=1}^N$ is ill-posed if the sample locations $\{t_i\}_{i=1}^N$ are unknown, even when $N \rightarrow \infty$ and the statistical distribution of those sample locations is known. However when accurate ordering information $t_1 < t_2 < ... < t_N$ is available, \cite{Kumar2015} derives asymptotic bounds for 1D signal reconstruction error when the locations of the (noisy) samples are drawn independently from a uniform distribution on $[0,1]$ (i.e., $\textrm{Uniform}(0,1)$). 
At its core, the derivation uses results regarding the expected value and variance of the order statistics of a uniform distribution. 
It is shown that the squared reconstruction error decays as $\bigo(1/N)$, and is proportional to $b^3$ and $\sigma^2$ where $b$ is the signal bandwidth and $\sigma$ is the noise standard deviation. 

\subsection{Classical Tomographic Reconstruction} 
\label{subsec:classical_tomo}
Tomography is a classical image reconstruction problem with applications in many areas, particularly medical imaging \cite{Gopal2020}. Consider an image $g$ defined on a domain $\Omega \subset \mathbb{R}^2$. Consider its Radon projection at angle $\theta$ given by
\begin{equation}
R_g(\rho,\theta) := \int \int_{\Omega}g(x,y) \delta(\rho - x \cos \theta - y \sin \theta) dx dy,
\label{eq:radon}
\end{equation}
where $\rho$ is called the translation or offset parameter, or bin index, and $\delta(.)$ is the Dirac delta function. For a fixed $\theta$, the signal $R_g(\rho,\theta)$ is a 1D signal. Given several such Radon projections at angles $\{\theta_i\}_{i=1}^N$ respectively, the aim is to reconstruct $g$. The basic principle is the well-known Fourier slice theorem (also known as the projection slice theorem) which states the following: The 1D Fourier transform of $R_g(\rho,\theta)$ with respect to $\rho$ and keeping $\theta$ fixed, denoted by $\mathcal{F}R^{\theta}_g$, is equal to a slice through the 2D Fourier transform of $g$ passing through the origin of the frequency plane at angle $\theta$. In other words, we have $\mathcal{F}R^{\theta}_g(\nu) = \mathcal{F}T_g(\nu \cos \theta, \nu \sin \theta)$, where $\mathcal{F}T_g$ stands for the 2D Fourier transform of $g$, and $\nu$ stands for frequency corresponding to the offset parameter $\rho$. The expression for $\mathcal{F}R^{\theta}_g$ is 
$\mathcal{F}R^{\theta}_g(\nu) = \int_{\mathbb{R}} R_g(\rho,\theta) e^{-j2\pi \nu \rho} d\rho$ where $j := \sqrt{-1}$. 
Likewise, the expression for $\mathcal{F}T_g$, and the corresponding expression for the inverse Fourier Transform in 2D, are given as follows:
\begin{equation}
    \mathcal{F}T_g(u,v) = \int\int_{\Omega} g(x,y) e^{-j2\pi(ux+vy)} dx dy, g(x,y) = \int\int \mathcal{F}T_g(u,v) e^{j2\pi(ux+vy)} dx dy.
\label{eq:FTf}
\end{equation}
where $u,v$ stand for frequency. Many approaches such as direct Fourier reconstruction~\cite{Stark1981} or filtered backprojection (FBP)~\cite{Kak2001}, reconstruct $g$ from its Radon projections. Reconstruction bounds for piecewise smooth images (belonging to Sobolev or Besov spaces) for different types of FBP filters are derived in \cite{Beckmann2015}. In classical tomography where the angles of projection are known accurately, it is well known that the reconstruction error decreases as the number of projection angles, spaced uniformly or drawn randomly from $\textrm{Uniform}(0,2\pi)$, is increased. The Fourier slice theorem here is presented in 2D, but can be easily extended to all higher dimensions including 3D.

\subsection{Tomographic Reconstruction from Unknown Angles of Projection}
In this paper, we are concerned with the problem of tomographic reconstruction when the angles of projection are unknown. This problem is primarily motivated by its occurrence in cryo-electron microscopy \cite{Frank2006} (cryo-EM). 
 In cryo-EM, a biologist places several hundred copies of a (3D) macromolecule suspended in a liquid medium on a slide. The slide along with the macromolecules is then frozen and placed under an electron microscope. The electron microscope produces an image called a micrograph containing hundreds of `particles' spread out on a background. Each particle is a 2D image corresponding to the parallel beam projection of a copy of the (3D) macromolecule. However, each copy is independently and randomly oriented and shifted, which effectively means that the precise projection direction is unknown. Cryo-EM reconstruction aims to determine the 3D structure of the macromolecule from the particles, and also estimate the orientations of the particles. 

In this work we consider 2D UVT, i.e. the 2D version of the problem where the structure to be determined is in 2D and its projections are 1D vectors. 2D UVT has been studied in prior art \cite{Coifman2008,Basu2000_a,Basu2000_b,Singer2013,Zehni2022}, as it is of interest from the point of view of setting up a theoretical foundation towards the more practical 3D problem. In some ways, the 2D problem is simpler as there are fewer degrees of freedom in specifying projection parameters. However in the 3D case, the Fourier slice theorem implies that the Fourier slices of projections from two different directions intersect at a `common line', which provides useful information for estimation \cite{Pragier2019}. Such common line information does not exist in 2D UVT. 
For 2D UVT, the uniqueness of the problem of joint estimation of projection angles and unknown image moments is proved in \cite{Basu2000_a}. The stability of this estimation problem given noise in the projections is established in \cite{Basu2000_b}. A numerical algorithm for the estimation task is proposed in \cite{Basu2000_b}, under the assumption that the angles are drawn from $\textrm{Uniform}(0,2\pi)$, but no theoretical analysis of the algorithm is presented. A procedure to estimate angular difference between projections given projection moments is detailed in \cite{Phan2017}. A method similar to the one in \cite{Basu2000_b} is developed in \cite{Coifman2008,Fang2011}, again assuming angles from $\textrm{Uniform}(0,2\pi)$. The projection vectors are first mapped down to $\mathbb{R}^2$ using a neighborhood distance preserving dimensionality reduction technique such as Laplacian Eigenmaps (LE) in \cite{Coifman2008} or the spherical locally linear embedding (SLLE) in \cite{Fang2011}. Then, the vectors in $\mathbb{R}^2$ are mapped to a `temporary' angle $\vartheta_i := \arctan \frac{y_i}{x_i}$ where $(x_i,y_i)$ is the 2D coordinate to which the $i$th projection vector was mapped via LE or SLLE. These temporary angles $\{\vartheta_i\}_{i=1}^N$ are calculated solely to produce a circular ordering of the projections. Arbitrarily assigning one of the projections to zero degrees, the other projections are assigned angles $\theta_i := \frac{2\pi i}{N}$, where $i \in [N]$, following the circular ordering obtained from $\{\vartheta_i\}_{i=1}^N$. This angle assignment is motivated by the fact that the mean value of the $i$th order statistic of a uniform distribution is $\frac{2\pi i}{N}$\textcolor{red}, and the variance of the order statistics lies between $\bigo(1/N^2)$ to $\bigo(1/N)$. In principle, this algorithmic approach is very similar to the one used in \cite{Kumar2015}, albeit for a very different application and without derivation of theoretical bounds. Given the angles $\{\theta_i\}_{i=1}^N$ obtained using order statistics of a uniform distribution, the image is reconstructed using FBP in \cite{Coifman2008,Fang2011}. A procedure to denoise projections prior to using them for reconstruction via the method from \cite{Coifman2008} is presented in \cite{Singer2013}. 

One advantage of the techniques proposed in \cite{Coifman2008,Basu2000_b,Fang2011} over the methods described in \cite{Scheres2012} is that the former are significantly faster, although they require knowledge of the angle distribution unlike the latter. This is because the latter techniques update the projection angles iteratively, whereas they are obtained in one shot in \cite{Coifman2008,Basu2000_b,Fang2011}. 
Most notably, the diverse 2D approaches in \cite{Basu2000_a,Basu2000_b,Coifman2008,Fang2011,Phan2017,Cheikh2017} as well as the 3D approaches in \cite{Scheres2012} do not present any formal theoretical analysis for the reconstruction error, which is the primary focus of this paper. More recently, \cite{Zehni2022} presents a generative adversarial network (GAN) based approach for 2D UVT reconstructions along with a method to determine the unknown angle distribution. The consistency of both estimates is established in the limit of infinitely many projections, but no analysis of sample complexity is provided, unlike our paper. 

\subsection{Overview of Approach in this Paper} 
The purpose of this paper is two-fold: (a) to extend the performance bounds for bandlimited signal reconstruction from unknown sample locations following a uniform distribution and known ordering to the case of \emph{quasi-bandlimited} (QBL) signal (cf. Sec.~\ref{subsec:qbl}) reconstruction from unknown sample locations with \emph{any known distribution} satisfying mild conditions and \emph{imperfectly} known ordering; and (b) to use this machinery to derive asymptotic error bounds to analyze the performance, i.e., expected squared reconstruction error, of an order statistics based algorithm for tomography with unknown angles from a known distribution. To support (b), an algorithm similar to the one developed in \cite{Coifman2008} (summarized in the earlier subsections) is followed in this paper, but using direct Fourier reconstruction instead of FBP (see Sec.~\ref{sec:experiments}). Recall that no theoretical analysis of the reconstruction error is presented in \cite{Coifman2008}. 
For (b), the inverse problem of interest is the reconstruction of 2D image $g$ from $N$ projection vectors $\{R_g(.,\theta_i)\}_{i=1}^N$, obtained at unknown projection angles $\{\theta_i\}_{i=1}^N$ drawn independently from some known distribution $F(.)$ which satisfies mild conditions such as strict monotonic increase, where its density function $f(.)$ is lower bounded by $\zeta > 0$ on the domain (satisfied by common distributions such as $\textrm{Uniform}(0,2\pi)$ and the von Mises distribution for a wide range of variance values). From the point of view of reconstruction, the samples available to us are the \emph{1D Fourier transforms} of each of the projection vectors, with respect to the bin index $\rho$. 

\begin{figure}
    \centering
    \includegraphics[width=0.35\textwidth]{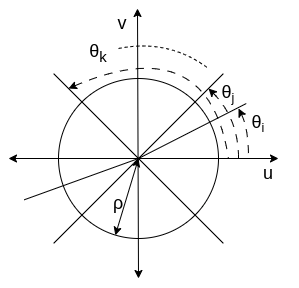}
    \caption{Diagram showing a ring for a constant $\rho$ but varying $\theta$ in the Fourier domain.}
    \label{fig:ft_spokes}
\end{figure}

Suppose, for the moment, that an ordering of the angles $\{\theta_i\}_{i=1}^N$ is known. By the Fourier slice theorem, the larger is the number of projection vectors at different angles, the larger is the number of radial lines in $\mathcal{F}T_g$ (all passing through the origin), defined in Eqn.~\eqref{eq:FTf}, that can be filled. Consider concentric rings in the 2D Fourier domain centered at $(0,0)$ for a fixed radius $\rho$ (equal to the offset parameter of the tomographic projection vector) across different angles. That is, each ring is a function $\mathcal{F}T_g(\rho \cos \theta, \rho \sin \theta)$ of angle $\theta$ for a fixed value of radius $\rho$ as shown in Fig.~\ref{fig:ft_spokes}.

Such rings are shown to be QBL (cf. Sec.~\ref{subsec:qbl}) in \cite{Pan1998} if the underlying image $g$ has bounded values. We then build upon the machinery from \cite{Kumar2015} for bandlimited signal reconstruction from USL following $\textrm{Uniform}(0,1)$. In doing so, we face certain challenges: 
(\textit{i}) We assume that the unknown projection angles are drawn independently either from $\textrm{Uniform}(0,2\pi)$ or from some \emph{other known distribution} satisfying mild conditions. We need to account for the possibly non-uniform nature of the underlying distribution of the sampling locations; 
(\textit{ii}) Unlike the case in \cite{Kumar2015} where the ordering of the samples was assumed to be known exactly, such an ordering must be \emph{estimated} for our application. Any such estimation procedure will have errors, which will have an effect on image reconstruction; (\textit{iii}) In our case, we cannot assume that the signal of interest, i.e., the Fourier ring described previously, is purely bandlimited, and have to allow for QBL signals (cf. Sec.~\ref{subsec:qbl}). This is also closely related to the fact that most images of interest are QBL and that their power spectrum decays rapidly beyond some `band limit'. 

In this paper, we systematically deal with these challenges to obtain the final set of theoretical performance bounds. 

\section{Signal Reconstruction in 1D from Samples at Unknown Locations}
\label{sec:reconstruction_1d}
\subsection{Original Problem}
\label{subsec:orig}
Consider a periodic, band-limited signal $g(t)$. 
We now define an important signal property which we refer to as property \textsf{L}:
\\
\textbf{Property} \textsf{L}: Without loss of generality, assume the period of $g(t)$ to be $1$ and $|g(t)|<1$ for all $t$ in the signal domain.
\\
Both assumptions in property \textsf{L} are reasonable, and the key results of this paper remain unchanged (scaled by constant factors) if the period of the signal is unequal to 1 or if its largest absolute value is unequal to 1. We present the following problem statement for band-limited signal reconstruction from unknown sample locations:

\noindent\textsf{(P0)} \textit{Consider an unknown signal $g(t)$ satisfying property \textsf{L}, and the (unknown) set $\{ t_1, t_2, ..., t_N\}$ which is created by drawing $N$ independent sample locations from $\textrm{Uniform}(0, 1)$ and then sorting them so that we have $t_1 < t_2 < ... < t_N$. Given $\{ g(t_1), g(t_2), ..., g(t_N)\}$ in this order and the bandwidth $b\in \mathbb{N}$ of $g(t)$, reconstruct $g(t)$.}

Since $g(t)$ is band-limited with a bandwidth $b$, we can express it in the form $g(t) = \sum_{k=-b}^{b} a_k e^{j2\pi kt}$ where $a_k \in \mathbb{C}$ is the $k$th Fourier series coefficient of $g(t)$ and $j := \sqrt{-1}$. Let $N$ be the number of samples. The following results are derived in \cite{Kumar2015}:
\begin{align}
    \label{eqn_gradient_bound}
    |g'(t)| &\leq 2\pi b,\\
    \label{eqn_orderstat_bound}
    N\xpcttn \left[ |t_i - \frac{i}{N}|^2 \right] &\leq \frac{1}{4} + \bigo(\sqrt{1/N}),\\
    \begin{split} \label{eqn_g_difference}
        \xpcttn [|g(t_i)-g(\frac{i}{N})|^2] &\leq \|g'(t)\|_\infty^2 \xpcttn \left[ |t_i - \frac{i}{N}|^2 \right]\leq (2\pi b)^2 \left( \frac{1}{4N} + \bigo \left( \frac{1}{N\sqrt{N}} \right) \right).
    \end{split}
\end{align}
Eqn.~(\ref{eqn_gradient_bound}) suggests that $g(t)$ is smooth. Eqn. (\ref{eqn_orderstat_bound}) states that for i.i.d. uniform samples, the expected squared difference between $i/N$ and the $i$th order statistic ($t_i$) decays with $N$. Eqn.~(\ref{eqn_g_difference}) follows from Lagrange's mean value theorem, and from Eqns.~(\ref{eqn_gradient_bound}) and (\ref{eqn_orderstat_bound}). 

By producing the following estimates $\hat{a}_k$ and $\hat{g}(t)$ of $a_k$ and $g(t)$ respectively:
\begin{align}
    \label{eqn_approx_defn}
    \hat{a}_k := \frac{1}{N} \sum_{i=1}^N g(t_i) \exp \left( -\frac{j2\pi ki}{N} \right), 
    \hat{g}(t) := \sum_{k=-b}^{b} \hat{a}_k e^{j2\pi kt},
\end{align}
the reconstruction method in \cite{Kumar2015} proves that:
\begin{equation}
    \label{eqn_asymp_closeness}
    \xpcttn [|\hat{a}_k - a_k|^2] \leq \frac{\pi^2 b^2}{N} + \bigo \left( \frac{1}{N\sqrt{N}} \right) + \frac{16\pi^2 b^2}{N^2}.
\end{equation}
Eqn.~(\ref{eqn_asymp_closeness}) suggests that the expected squared difference between the estimated Fourier series coefficient $\hat{a}_k$ and the actual coefficient $a_k$ asymptotically goes to $0$ as the number of samples goes to infinity. Parseval's theorem is then invoked to show that
\begin{equation}
    \label{eqn_bandlimited_recon_ub}
   \xpcttn \left[ \|\hat{g} - g\|_2^2 \right] = \sum_{k=-b}^b \xpcttn \left[ |\hat{a}_k - a_k|^2 \right] = \bigo \left( \frac{b^3}{N} \right) \xrightarrow[N \uparrow \infty]{} 0,
\end{equation}
ignoring the $\bigo(1/N^2)$ and $\bigo(1/N\sqrt{N})$ terms as they are dominated by the $\bigo(1/N)$ term.

\subsection{Quasi-bandlimited signals}
\label{subsec:qbl}
Consider any periodic signal $g(t)$ satisfying $|g(t)| < 1$. Using the Fourier series representation we can write
$g(t) = \sum_{k=-\infty}^{\infty} a_k e^{j2\pi kt},
$ for some Fourier series coefficient $a_k \in \mathbb{C}$. We say that this signal is quasi-bandlimited (QBL) with parameters $k_1 \in \mathbb{N}, \gamma \in \mathbb{R}_{+}$, denoted as $g \sim \textrm{qbl}(k_1, \gamma)$, if
$\exists \ d \geq 0 \ s.t. \ |a_k| \leq de^{-\gamma|k|} \ \forall \ k \in \mathbb{Z}: |k| \geq k_1$. Note that this is a definition for QBL signals that we have presented to facilitate our analysis. The definition, however, is obeyed by a broad class of functions including the Fourier transform of the Radon transform of images with bounded value, as discussed in \cite{Pan1998}. Intuitively, the Fourier series coefficients of a QBL signal decay rapidly in magnitude beyond a certain frequency limit, here denoted by $k_1$. The property of band-limitedness and quasi-bandlimitedness are together analogous to the concepts of sparsity and weak sparsity in the CS or sparse regression literature -- the former implies that a small number of signal coefficients are non-zero, whereas the others are exactly equal to zero; the latter implies that a small number of signal coefficients are sufficiently large in magnitude, whereas the others are very small in magnitude (close to zero, or possessing an exponential decay). We now state the problem of QBL signal reconstruction from unknown samples: 

\noindent\textsf{(P1)} \textit{Consider an unknown QBL $g \sim \textrm{qbl}(k_1, \gamma)$ having period 1 with $\forall t, |g(t)| < 1$ and the (unknown) set $\{ t_1, t_2, ..., t_N\}$ which is created by drawing $N$ sample locations independently from an invertible distribution $F(.)$, with corresponding probability density function $f(.)$, defined on the domain $[0,1]$, and then sorting them so that $t_1 < t_2 < ... < t_N$. We consider $f(.)$ to be lower bounded by $\zeta > 0$ everywhere on $[0,1]$. Given $\{ g(t_1), g(t_2), ..., g(t_N)\}$ in this order and the parameters $k_1, \gamma$ of $g(t)$, reconstruct $g(t)$.}

\noindent\textbf{Remark:} The requirement of $f(.) \geq \zeta > 0$ is reasonable, as it merely emphasizes the fact that if no samples were drawn from some region of $[0,1]$, we would not be able to reconstruct the signal correctly in that region. Common distributions such as $\textrm{Uniform}(0,1)$ and a truncated Gaussian defined on $[0,1]$ respect this lower bound on $f(.)$. 

\subsection{Model for Errors in Specification of Sample Ordering}
\label{subsec:models_err_ordering}
\begin{figure}
    \centering
    \includegraphics[width=0.4\linewidth]{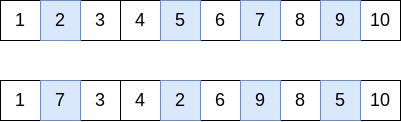}
    \includegraphics[width=0.4\linewidth]{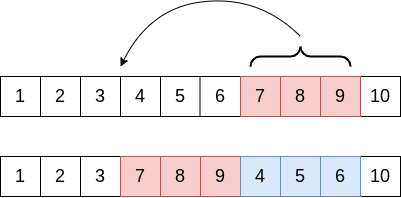}
    \caption{An example of shuffles (left two rows) and shifts (right two rows).}
    \label{fig:shift_shuffle_eg}
\end{figure}

We pronounce a bijection $h: [N] \rightarrow [N]$ to be `good' with parameters $\bar{\delta}, N_{\bar{\delta}} \in \mathbb{N}$ (denoted as $h \sim \textrm{good}(\bar{\delta}, N_{\bar{\delta}})$) if $\sum_{i=1}^N \indctr \{ |i - h(i)| > \bar{\delta} \} \leq N_{\bar{\delta}}$. That is, at most $N_{\bar{\delta}}$ mappings obtained via $h(.)$ are more than $\bar{\delta}$ units away from their original values. Note that $h \sim \textrm{good}(\bar{\delta},N_{\bar{\delta}})$ covers errors such as the following: (\textit{i}) \textbf{Shuffles} - a random subset of the sample locations are shuffled; (\textit{ii}) \textbf{Shifts} - a block of consecutive samples is chosen at random and inserted into a random position, displacing all samples after this position. We specifically consider shuffles and shifts, because we have seen in our experiments that these errors occur commonly in any assignment operation, as will be shown in Sec.~\ref{sec:experiments}. Refer to Fig.~\ref{fig:shift_shuffle_eg} for some pictorial examples of shuffles and shifts. 

\noindent\textsf{(P2)} 
\textit{Consider all the conditions on signal $g$ and sampling locations $t_1, t_2,...t_N$ as defined in \textsf{P1}. In addition, consider some unknown bijective map $h \sim \textrm{good}(\bar{\delta}, N_{\bar{\delta}})$. The set $\{ t'_1, t'_2, ..., t'_N\}$ is the permutation of $\{ t_1, t_2, ..., t_N\}$ by the map $h$ (i.e., $t'_i = t_{h(i)}$). Given $\{ g(t'_1), g(t'_2), ..., g(t'_N)\}$ in this order, the parameters $k_1, \gamma$ of $g(t)$ and the parameters $\bar{\delta}, N_{\bar{\delta}}$ of $h(\cdot)$, reconstruct $g(t)$.}

\subsection{Model for Noise in Samples}
\label{subsec:noise_in_samples}
Consider a noise process $\varepsilon(t)$ which has the following properties: (\textit{i}) at each point $t$, $\varepsilon(t)$ has mean $0$ and variance $\sigma^2$; (\textit{ii}) $\varepsilon(t)$ does not depend on the signal $g(t)$ or on the set of sampling locations $\{ t_1, t_2, ..., t_N\}$; (\textit{iii}) for any $N \in \mathbb{N}$ and distinct $\{ t_1, t_2, ..., t_N\}$, the noise values $\{\varepsilon(t_1), \varepsilon(t_2), ..., \varepsilon(t_N)\}$ are statistically independent.
We now state the problem of QBL signal reconstruction from noisy samples at unknown locations with errors in specification of sample ordering:

\noindent\textsf{(P3)} 
\textit{Consider all the conditions on signal $g$ and sampling locations $t_1, t_2,\ldots,t_N$ as defined in \textsf{P2}. Given $\{ g(t'_1)+\varepsilon(t'_1), g(t'_2)+\varepsilon(t'_2),\ldots, g(t'_N)+\varepsilon(t'_N)\}$ in this order, the parameters $k_1, \gamma$ of $g(t)$ and the parameters $\bar{\delta}, N_{\bar{\delta}}$ of $h(\cdot)$, reconstruct $g(t)$.}

\subsection{1D signal reconstruction with ordering errors}
\label{subsec:reconstruction_1d_ordering_errors}
Given a distribution $F$ of the sampling location, let $x_i$ denote its $i^{th}$ order statistic. We note that $F(x_i) \sim$ $\textrm{Uniform}(0,1)$. Hence if we draw a sample $X$ from $\textrm{Uniform}(0,1)$, then $F^{-1}(X)$ acts a sample from $F(.)$. We use this fact in our first result, which is stated below. The result is proved in the supplemental material.
\begin{theorem}
\label{thm:1d_reconstruction}
Consider the setting in \textsf{(P3)}. Choose $k_0 = \left\lceil \frac{\log(N)}{\gamma} \right\rceil$ as the bandwidth of a band-limited approximation, $g^{k_0}(t)$, of the original QBL signal $g(t)$ (i.e., $g^{k_0}(t) := \sum_{|k| \leq k_0} a_k e^{j2\pi kt}$). Define our reconstructed signal as follows: 
\begin{align}
    \label{eqn_recon_coeff_defn}
    \Bar{a}_k &:= \sum_{i=1}^N (g(t'_i)+\varepsilon(t'_i))(x_i-x_{i-1}) \exp \left( -j2\pi k x_i \right),    \Bar{g}^{k_0}(t) &:= \sum_{k=-k_0}^{k_0} \Bar{a}_k e^{j2\pi kt},
\end{align}
where $\{t'_i\}_{i=1}^N$ are as defined in (\textsf{P3}), $\{\Bar{a}_k\}_{k=1}^N$ are discrete approximations of the reconstructed Fourier Series coefficients given $\{g(t'_i) + \varepsilon(t'_i)\}_{i=1}^N$, $\Bar{g}^{k_0}(t)$ is the reconstructed signal with bandwidth $k_0$ and Fourier Series coefficients $\Bar{a}_k$, and $x_i := F^{-1}(i/N)$ where $f(.), F(.)$ are the known PDF and CDF of the (unknown) sampling locations respectively with $F(.)$ assumed to be invertible everywhere. Then, for $N > e^{\gamma k_1}$,
\begin{align}
    &\xpcttnb{\|g(t) - \Bar{g}^{k_0}(t)\|_2^2} = 
    \bigob{\frac{\sigma^2 k_0}{N\zeta^2}} + \bigob{\frac{N_{\bar{\delta}}k_0}{N \zeta^2}} + \bigob{\frac{k_0}{N^2 \zeta^2}} + \bigob{\frac{k^3_0}{N \zeta^4} + \frac{k^3_0 \bar{\delta}^2}{N^2 \zeta^4}} \nonumber\\
    &\qquad= \bigob{\frac{(\sigma^2 + N_{\bar{\delta}})\log N}{N \gamma \zeta^2} + \frac{\log^3 N}{N \gamma \zeta^4} + \frac{\bar{\delta}^2 \log^3 N}{N^2 \gamma^3 \zeta^4}}.
\end{align}
\end{theorem}
\noindent\textbf{Remarks:}
(\textit{i}) \textbf{Expression for $\Bar{a}_k$:} The Fourier series coefficients are obtained using an inner product of the form $\langle g(t) + \varepsilon(t), \exp(-j2\pi kt) \rangle := \int_{0}^{1} (g(t) + \varepsilon(t)) \exp(-j2\pi kt) dt$, which is approximated using a Riemann sum with possibly unequal intervals of the form $x_i - x_{i-1}$ (in contrast to Eqn.~\ref{eqn_approx_defn} where the intervals were equal). Note that the different $x_i$ values correspond to those locations for which $F(x_i) = i/N$. In case of $\textrm{Uniform}(0,1)$, we see that $x_i = i/N$.
    (\textit{ii}) \textbf{Choice of $k_0$:} In practice, the exact value of $k_1$ for the QBL signal is unknown, and $k_0$ is intended to act as an upper bound on $k_1$. The choice of $k_0 = \left\lceil \frac{\log(N)}{\gamma} \right\rceil$ ensures that the residual error obtained by ignoring frequency components with $|k| > k_0$ due to the QBL nature of the signal, decreases with $N$. This is clear from Lemma 5 stated and proved in supplemental material. 
    (\textit{iii}) \textbf{Statistical Consistency:} Given a constant $\sigma$ that does not depend on $N$, statistical consistency for the reconstruction i.e. reconstruction error tending towards 0 as $N \rightarrow \infty$, is guaranteed if $N_{\bar{\delta}} \log(N)$ is $o(N)$ and $\bar{\delta}^2 \log^3 N$ is $o(N^2)$.
    (\textit{iv}) Within a bandwidth of $k_0$, the reconstructed Fourier Series coefficients, $\Bar{a}_k$, are `close' to the actual coefficients $a_k$. How `close' the reconstructed coefficients are, depends on how `good' the bijective map $h$ is, i.e., how small are the values $\bar{\delta}, N_{\bar{\delta}}$, and how small is the noise standard deviation $\sigma$. Also see Lemma~\ref{lem:fr_ring_reconstruction}.
    (\textit{v}) Setting $\varepsilon(t'_i)=0$ in Eqn.~\ref{eqn_recon_coeff_defn} produces a reconstruction for the case when $\sigma=0$, which is the same as \textsf{(P2)}. The expected reconstruction error then is $\bigob{\frac{N_{\bar{\delta}}k_0}{N \zeta^2}} + \bigob{\frac{k_0}{N^2 \zeta^2}} + \bigob{\frac{k^3_0}{N \zeta^4} + \frac{k^3_0 \bar{\delta}^2}{N^2 \zeta^4}}$.
    (\textit{vi}) For the case when $f(.)$ is $\textrm{Uniform}(0,1)$, we set $\zeta = 1$ to obtain the upper bound $    \bigob{\frac{\sigma^2 k_0}{N}} + \bigob{\frac{N_{\bar{\delta}}k_0}{N}} + \bigob{\frac{k^3_0}{N} + \frac{k^3_0 \bar{\delta}^2}{N^2}}$. Moreover, using $g(t_i)$ in place of $g(t'_i)$ in Eqn.~\ref{eqn_recon_coeff_defn} produces a reconstruction for the case when $\bar{\delta}=N_{\bar{\delta}}=0$, which is the same as \textsf{(P1)}. The expected reconstruction error in such a case is $\bigob{\frac{k_0^3}{N}} + \bigob{\frac{\sigma^2 k_0}{N}}$. The bounds exactly match with the result in \cite{Kumar2015} for \textsf{(P0)}, stated in Eqn.\eqref{eqn_bandlimited_recon_ub}, when we replace $k_0$ with the true bandwidth $b$. For non-uniform $f$, it is clear that the worst case reconstruction is better for larger values of $\zeta$, the lower bound on the density. 

This result is a strict generalization of that in \cite{Kumar2015} as it allows for errors in the specified ordering, considers QBL signals instead of strictly bandlimited signals, and considers non-uniform sampling distributions. The proof of this theorem in the supplemental material essentially bounds four types of errors: error due to bandlimited approximation of a quasibandlimited signal (cf. Lemma 5 of supplemental material), error due to reconstruction from noisy samples (cf. Lemma 8 of supplemental material), error due to mis-specification in the sample location ordering (cf. Lemmas 9-12 of supplemental material), and the error due to reconstruction from samples at unknown locations but using order statistics information (cf. Lemmas 13-16 of supplemental material). We summarize the difference in the assumptions made by our work as compared to \cite{Kumar2015}, as well as to our group's previous work on related problems from \cite{Pandotra2019,Vaish2022}, in Table~\ref{tab:comparison}.
\begin{table*}[h]
    \footnotesize
    \centering
    \begin{tabular}{|c|c|c|c|c|}
    \hline
    Property & \cite{Kumar2015} & \cite{Pandotra2019} & \cite{Vaish2022} & This work \\\hline
    Sample location & \ding{56} & \checkmark? & \checkmark? & \ding{56} \\
    Sample ordering & \checkmark & NN & NN & \checkmark?\\
    Sample distribution & \checkmark, U & NN & NN & \checkmark, NU\\
    Can handle sample noise & \checkmark & \checkmark & \checkmark & \checkmark\\
    Signal Property & BL & Wavelet-sparse & Wavelet-sparse & QBL\\
    Application & Mobile sensors & MRI & DMRI & 2D UVT\\\hline
    \end{tabular}
    \caption{Comparisons between assumptions made by different techniques: In this table, \checkmark means `known perfectly', \ding{56} means `unknown', \checkmark? means `known imprecisely' and NN means `Not needed'; BL means bandlimited and QBL means quasi-bandlimited (both in Fourier sense); `U' refers to $\text{Uniform}(0,1)$ and `NU' refers to possibly non-uniform distribution. In \cite{Vaish2022}, \cite{Pandotra2019}, an approximate knowledge of the sample locations is required and available in the applications they deal with. In our work and \cite{Kumar2015}, the applications dealt with do not have access to sample locations at all.}
    \label{tab:comparison}
\end{table*}

\subsection{Comparison with Compressed Sensing}
\label{sec:CS_comparison}
In the presented technique, the sample values are available, but the sampling locations are completely unknown. Only the sample ordering and the sample location distribution are known. Given the ordering, the concept of order statistics is used to obtain the locations. In order to use this concept, it is \emph{necessary} to have a large number of samples ($N$), otherwise the variance of the estimate (between $O(1/N^2)$ and $O(1/N)$) will be too large. This situation stands in stark contrast to compressed sensing, which uses a \emph{small} number of samples at \emph{known} locations. Our group has previously explored compressed sensing in the case of \emph{uncertain} locations \cite{Pandotra2019,Vaish2022}, i.e. locations with some (small) error in their specification. However in the present work, the locations are completely unknown. As such, the CS technique also stands in stark contrast to the 2D UVT problem considered in this paper. The technique in this paper uses the QBL property of the underlying signal, which is a special form of approximate signal property. Using a more general form of (approximate) sparsity, that is commonly employed in CS, within the current framework is challenging and may not even be required.

\section{Tomographic Image Reconstruction from Unknown Projection Angles}
\label{sec:tomog}
Consider a 2D signal $g(x, y)$ that satisfies the following properties: (\textsf{L1}) $g$ has compact support on a closed, bounded domain $\Omega \subset \mathbb{R}^2$, i.e., $\exists \ r_0 \ s.t. \ g(x, y) = 0 \ \forall (x, y) : x^2 + y^2 > r_0^2$); (\textsf{L2}) $g$ is real-valued and bounded, i.e., $\exists \ a_m \ s.t. \ |g(x, y)| \leq a_m \forall (x, y) \in \Omega$.

Refer to Sec.~\ref{subsec:classical_tomo} for expressions for the Radon transform of $g$ (i.e., $R_g$), for the Fourier slice theorem, and for the expression for the Fourier transform of $R_g$ respectively. Given the statement of the Fourier slice theorem, one can conclude that the constant-$\nu$ slice in $\mathcal{F}R_g$ varying across $\theta$ is the same as an origin-centred ring of the same radius in $\mathcal{F}T_g$ (refer Fig. SM.1 of the supplemental material). We now formally state the problem of 2D tomography under unknown angles. 

\noindent \textsf{(P4)} \textit{Consider some unknown 2D signal $g$ satisfying the aforementioned properties (\textsf{L1}) and (\textsf{L2}). Consider the ordered set $A := \{ \theta_1, \theta_2, \cdots, \theta_N \}$ which is created by drawing $N$ angles independently from some invertible probability distribution $F(.)$ on $[0, 2\pi)$ whose density $f(.)$ is lower bounded by some $\zeta > 0$, and then sorting them. Furthermore, consider the set $B := \{ \theta_{\tilde{h}(1)}, \theta_{\tilde{h}(2)}, \cdots \theta_{\tilde{h}(N)} \}$ which is a permutation of $A$ by the (unknown) bijection $\tilde{h}$ defined in Sec.~\ref{subsec:models_err_ordering}. Also consider noise values $\epsilon_1(\rho), \epsilon_2(\rho), \cdots, \epsilon_N(\rho)$ such that for all $i, \rho$, $\epsilon_i(\rho)$ is drawn independently from $N(0,\sigma^2)$. Given the set $Q(.) := \{R_g(\cdot, \theta_{\tilde{h}(1)}) + \epsilon_1(\cdot), R_g(\cdot, \theta_{\tilde{h}(2)}) + \epsilon_2(\cdot), \cdots ,R_g(\cdot, \theta_{\tilde{h}(N)}) + \epsilon_N(\cdot) \}$, reconstruct $g$.}

In the rest of this section, we state three lemmas, a corollary and a theorem, all of which are proved in the supplementary material. Note that in UVT, it is possible to reconstruct an image only up to a rotational shift \cite{Basu2000_a}, because the ordering between projections can only be found relative to other projections. Hence it is impossible to order them with respect to any fixed axis. This is also true for a few other problems such as structure from motion in computer vision \cite{Trucco1998}. For brevity in our analysis, we assume that $\theta_1$ is known and equal to 0, and $\tilde{h}(1) = 1$, without loss of generality.

\begin{lemma}
    \label{lem:fr_ring_is_qbl}
    Given image $g$ satisfying properties (\textsf{L1}) and (\textsf{L2}), we have $
        \mathcal{F}R^{\theta}_g(\nu_1) \sim \textrm{qbl}(k_1,\gamma)\textrm{ where } k_1 :=\frac{2\pi \nu_1 r_0}{0.765}, \gamma:=0.765$. That is, $\mathcal{F}R^{\theta}_g(\nu_1)$ is a QBL signal in $\theta$. 
\end{lemma}
The factor $0.765$ is due to properties of the Bessel functions as shown in \cite[Appendix A]{Pan1998}. Even though the ordering of projection angles is unknown, it can be estimated if the distribution $F(.)$ of the angles is known, by using the reasonable assumption that projections from nearby angles are structurally similar. Hence the distance between the projections (in an appropriate feature space) increases as the angular difference between them increases. This intuition is used in the Laplacian eigenmaps technique in \cite{Coifman2008} to derive the angle ordering, following by angle assignment using order statistics of $F(.)$. 
Let the true ordering of projections be $P(\cdot) := \{ R_g(\cdot, \theta_1), R_g(\cdot, \theta_2), \cdots, R_g(\cdot, \theta_N) \}$ with $0 = \theta_1 < \theta_2 < \cdots < \theta_N < 2\pi$. Let the ordering of projections produced by some algorithm (from the input $Q$) be $P^{\prime}(.) := \{ R_g(\cdot, \theta_{h(1)})+\epsilon_1(\cdot), R_g(\cdot, \theta_{h(2)})+\epsilon_2(\cdot), \cdots, R_g(\cdot, \theta_{h(N)}+\epsilon_N(\cdot)) \}$. $P'$ is characterized by the mapping $h$, which could be unequal to the mapping $\tilde{h}$ used for defining $Q$. Since it is assumed that $q_1 = R_g(\cdot, \theta_1=0)$, there is no rotational ambiguity, and the algorithm should also set $h(1) = 1$.

\begin{lemma}
\label{lem:fr_ring_reconstruction}
Given the set of re-ordered projections $P^{\prime}(.) = \{ R_g(\cdot, \theta_{h(1)})+\epsilon_1(\cdot), R_g(\cdot, \theta_{h(2)})+\epsilon_2(\cdot), \cdots, R_g(\cdot, \theta_{h(N)})+\epsilon_N(\cdot) \}$ of 2D signal $g$, such that $h \sim \textrm{good}(\bar{\delta}, N_{\bar{\delta}})$, set $k_0 := \left\lceil \frac{\log N}{0.765} \right\rceil$. For any $\nu_1 > 0$, define
$a_k := \frac{1}{N} \sum_{i=1}^N \left(\mathcal{F}R^{\theta_{h(i)}}_g(\nu_1)+\epsilon_i(\nu_1)\right) (x_i-x_{i-1}) e^{ -j k x_i },  \widehat{\mathcal{F}R}^{\theta}_g(\nu_1) := \sum_{k=-k_0}^{k_0} a_k e^{j k\theta}$, where $x_i := 2\pi F^{-1}(i/N)$. Then, for $N > e^{2\pi \nu_1 r_0}$,
\begin{align}
\xpcttnb{\left\|\mathcal{F}R^{\theta}_g(\nu_1) - \widehat{\mathcal{F}R}^{\theta}_g(\nu_1)  \right\|_2^2} =  \bigob{\frac{\sigma^2 k_0}{N\zeta^2}} + \bigob{\frac{N_{\bar{\delta}}k_0}{N \zeta^2}} + \bigob{\frac{k_0}{N^2 \zeta^2}} + \bigob{\frac{k^3_0}{N \zeta^4} + \frac{k^3_0 \bar{\delta}^2}{N^2 \zeta^4}} \nonumber \\
    = \bigob{\frac{(\sigma^2 + N_{\bar{\delta}})\log N}{N \zeta^2} + \frac{\log^3 N}{N \zeta^4} + \frac{\bar{\delta}^2 \log^3 N }{N^2 \zeta^4}}.
    \end{align}
\end{lemma}
\vspace{-0.6in}
\begin{corollary}
\label{cor:disc_recon}
For $N > e^{2\pi \nu_0 r_0}$ and $h \sim \textrm{good}(\bar{\delta}, N_{\bar{\delta}})$,
\begin{align}
 &\int_0^{\nu_0} \xpcttnb{\left\| \mathcal{F}R^{\theta}_g(\nu) - \widehat{\mathcal{F}R}^{\theta}_g(\nu)\right\|_2^2} d\nu = \bigob{\frac{\nu_0 \sigma^2 k_0}{N\zeta^2}} + \bigob{\frac{\nu_0 N_{\bar{\delta}}k_0}{N \zeta^2}} + \bigob{\frac{\nu_0 k_0}{N^2 \zeta^2}} \nonumber \\
 &\qquad\qquad+ \bigob{\frac{\nu_0 k^3_0}{N \zeta^4} + \frac{\nu_0 k^3_0 \bar{\delta}^2}{N^2 \zeta^4}} =\bigo \Bigg(\frac{\nu_0(\sigma^2 + N_{\bar{\delta}})\log N}{N \zeta^2} + \frac{\nu_0 \log^3 N}{N \zeta^4} + \frac{\nu_0 \log^3 N \bar{\delta}^2}{N^2 \zeta^4} \Bigg). 
\end{align}
\end{corollary}
Note that reconstructing a ring of $\mathcal{F}R^{\theta}_g(\nu_1)$ is equivalent to reconstructing a ring of $\mathcal{F}T_g$ with radius $\nu_1$ (as discussed previously), and hence if we perform the reconstruction in Lemma~\ref{lem:fr_ring_reconstruction} for every $\nu_1 \leq \nu_0$ (where $\nu_0$ is an arbitrarily chosen radius), we can reconstruct the disc of radius $\nu_0$ in $\mathcal{F}T_g$. Thus our reconstruction can be defined as $\hat{g} := \mathcal{F}^{-1}(\widehat{\mathcal{F}T}_g)$ where:
\begin{align}
     \widehat{\mathcal{F}T}_g(u,v) &:= \begin{cases}
     \widehat{\mathcal{F}R}^{\theta}_g(\nu=\sqrt{u^2+v^2});         u^2+v^2 \leq \nu_0^2, \theta=\tan^{-1}(\frac{v}{u})\\
        0; \textrm{otherwise}.
     \end{cases}
     \label{eqn_frecon_defn}
\end{align}
Here $\mathcal{F}^{-1}$ stands for inverse Fourier Transform. Since we are considering bandlimited approximation of a QBL signal, we now present a result to bound the tail of the Fourier transform magnitude outside the radius $\nu_0$ for an image which obeys a certain kind of power law as described in \cite{Beckmann2015}. 
\begin{lemma}
    \label{lem:sobolev}
    Consider $\alpha \in \mathbb{R}_{+} \cup \{0\}$ and $g:\Omega \subset \mathbb{R}^2 \rightarrow \mathbb{R}$. Define $\|g\|_\alpha^2 := \frac{1}{2\pi} \int\int (1+u^2+v^2)^\alpha |\mathcal{F}T_g(u, v)|^2 du dv$. If $\|g\|_\alpha^2 < \infty$,
    then for any $\nu_0>0$ the tail of the magnitude of $\mathcal{F}T_g$ outside the radius $\nu_0$ is bounded by $e_g(\nu_0) := \frac{1}{2\pi} \int_{u^2+v^2>\nu_0^2} |\mathcal{F}T_g(u, v)|^2 dv du \leq \nu_0^{-2\alpha} \|g\|_\alpha^2$.
\end{lemma}
\begin{theorem}
    Consider some arbitrarily chosen radius $\nu_0$. For $N > e^{2\pi \nu_0 r_0}$, $k_0 := \left\lceil \frac{\log N}{0.765} \right\rceil$, $g$ which satisfies the conditions in Lemma~\ref{lem:sobolev}, and the reconstruction $\hat{g}$ defined in Eqn.~\ref{eqn_frecon_defn}, we have
\begin{align}
\xpcttn \left[ \| g - \hat{g} \|^2 \right] = \bigob{\frac{\nu_0 \sigma^2 k_0}{N\zeta^2}} + \bigob{\frac{\nu_0 N_{\bar{\delta}}k_0}{N \zeta^2}}  + \bigob{\frac{\nu_0 k_0}{N^2 \zeta^2}} + \bigob{\frac{\nu_0 k^3_0}{N \zeta^4} + \frac{\nu_0 k^3_0 \bar{\delta}^2}{N^2 \zeta^4}} + \nonumber \\  \bigob{\nu_0^{-2\alpha}\|g\|_\alpha^2}\nonumber 
=\bigob{\frac{\nu_0(\sigma^2 + N_{\bar{\delta}})\log N}{N \zeta^2} + \frac{\nu_0 \log^3 N}{N \zeta^4} + \frac{\nu_0 \log^3 N \bar{\delta}^2}{N^2 \zeta^4}} + \bigob{\nu_0^{-2\alpha}\|g\|_\alpha^2}. 
\end{align}
\label{thm:2d_reconstruction}
\end{theorem}
\vspace{-0.7in}
This theorem has the following implications:
(\textit{i}) \textbf{Choice of $\nu_0$}: In practice, $\nu_0$ can be selected based on statistics of a class of similar images observed earlier. It can also be chosen by observing the 1D Fourier transforms of the individual Radon projections. As such, $\nu_0$
    should be chosen to balance the tradeoff between the two terms of Thm.~\ref{thm:2d_reconstruction}. A larger $\nu_0$ implies that a larger disc in the Fourier space is reconstructed, and hence the residual error (second term) is lower. However, a larger $\nu_0$ directly increases the reconstruction error within the disc, owing to a larger number of coefficients that need to be estimated.   (\textit{ii}) \textbf{Statistical consistency} (i.e., the reconstruction error $\rightarrow 0$  as $N\rightarrow \infty$) is guaranteed if the recovered order of the projections characterized by the mapping function $h$ is ``good" with $\bar{\delta} = o\left(\frac{N}{(\log N)^{3/2}}\right), N_{\bar{\delta}} = o\left(\frac{N}{\log N}\right)$ and $\nu^2_0$ is $o(N)$. Our experiments in Sec.~\ref{sec:experiments} reveal that the assumption of a `good' mapping as output from a tractable algorithm is reasonable. (\textit{iii}) \textbf{Effect of noise:} For fixed $\bar{\delta}, N_{\bar{\delta}}$, the effect of noise in the projections diminishes with the number of projections $N$. In practice however, a large amount of noise might adversely affect the algorithm that orders the projections. Hence, the projections need to be denoised \cite{Singer2013,Malhotra2016} prior to recovering their ordering. 
    (\textit{iv}) When the distribution function $F$ is $\textrm{Uniform}(0,2\pi)$, we have $\zeta=1/2 \pi$, which allows for tighter upper bounds.

While Thm.~\ref{thm:2d_reconstruction} is a theoretical result, its idea can be adapted into the following reconstruction algorithm, which we call \textsf{RA}. A detailed pseudo-code for the algorithm is given in Alg. SM.1 of supplemental material.
(1) Given a set of $N$ noisy projections drawn from distribution $F(.)$, denoise them using a PCA-based technique \cite{Singer2013}, and then order them using the Laplacian eigenmaps (LE) technique from \cite{Coifman2008}. Assign angle values to each of the projections based on order statistics of the appropriate sampling distribution to the $i$th projection.
(2) Compute the 1D Fourier Transform of each projection using an oversampled DFT. For each different radius value $\nu \leq \nu_0$ of the DFT, reconstruct the corresponding coefficient $a_k$ using Lemma~\ref{lem:fr_ring_reconstruction}. The number of coefficients, $k_0$, is a problem-dependent choice, and we found $k_0 = \bigob{\log N}$ to work well, as stated in the main theorems. Note that for a fixed $\nu$, a ring of coefficients is formed in the Fourier domain.
(3) Choose a discretization $M$ such that the Fourier space for angle $\theta$ while reconstructing $\widehat{\mathcal{F}R}_g(\nu_1,\theta)$ in Lemma~\ref{lem:fr_ring_reconstruction} is densely covered by $M$ uniformly spaced spokes. Using the previously computed coefficients $a_k$, reconstruct the value of each Fourier ring at $M$ uniformly spaced points in $[0, 2\pi)$ using the expression for $\widehat{\mathcal{F}R}^{\theta}_g(\nu_1)$ in Lemma~\ref{lem:fr_ring_reconstruction}.
(4) Perform an inverse polar Fourier transform, say using a library for a non-uniform fast Fourier transform, such as pyNUFFT \footnote{\url{https://jyhmiinlin.github.io/pynufft/index.html}}. This yields the final reconstructed image. 

\section{Experiments}
\label{sec:experiments}
Here, we present numerical experiments with an aim to verify theoretical results such as Thm.~\ref{thm:2d_reconstruction}. We do not aim to compare with algorithms such as \cite{Malhotra2016,Basu2000_b,Zehni2022}, as the primary aim of this paper is to provide analytical bounds, which do not exist for these aforementioned approaches. 

Our experiments were performed using a 2D image of a ribosome (size $608 \times 608$, one pixel width is 2.78 Angstrom) from the well-known Electron Microscopy Data Bank (EMDB)\footnote{EMD-26959 Single-particle, \url{https://www.ebi.ac.uk/emdb/EMD-26959}}. Given this image, $N$ 1D Radon projections were computed at angles drawn independently from a von Mises distribution with density given by $f(\theta;\mu,\kappa) = e^{\kappa\cos(\theta-\mu)}/(2\pi I_0(\kappa))$ (referred to henceforth as $\textrm{VM}(\mu,\kappa)$) with mean $\mu = \pi$ or inverse variance $\kappa \in \{0,1,2\}$ (higher $\kappa$ indicates peakier density and $\kappa = 0$ yields $\textrm{Uniform}(0,2\pi)$). Zero mean Gaussian noise with standard deviation $\sigma = f_n \times$ the average noiseless projection values, was added to every element of each projection (where $f_n \in [0,1]$) followed by PCA based denoising. The aim was to reconstruct the image without knowledge of these angles, using algorithm \textsf{RA} described at the end of Sec.~\ref{sec:tomog}, mainly in order to verify Thm.~\ref{thm:2d_reconstruction}. 

\begin{figure}[htb]
    \centering
     \includegraphics[width=0.48\textwidth]{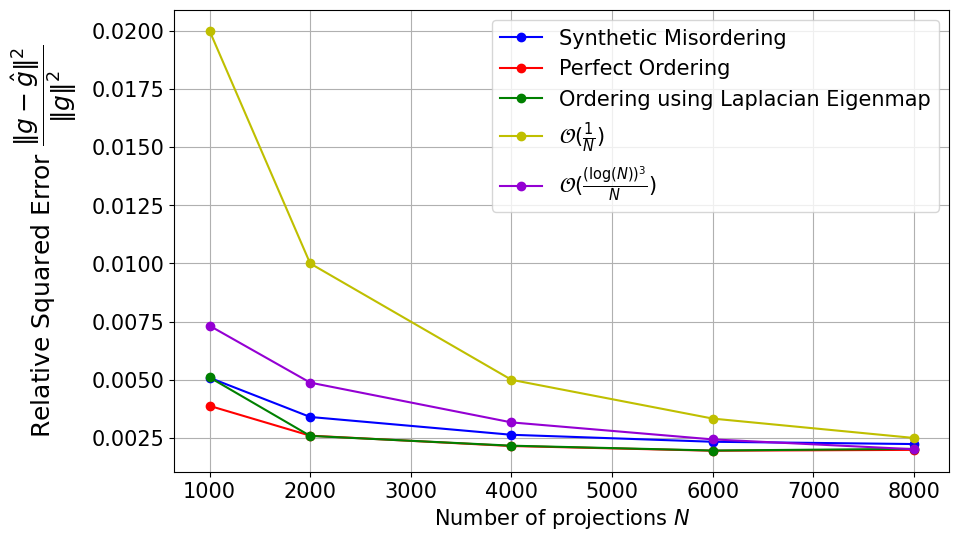}   \includegraphics[width=0.48\textwidth]{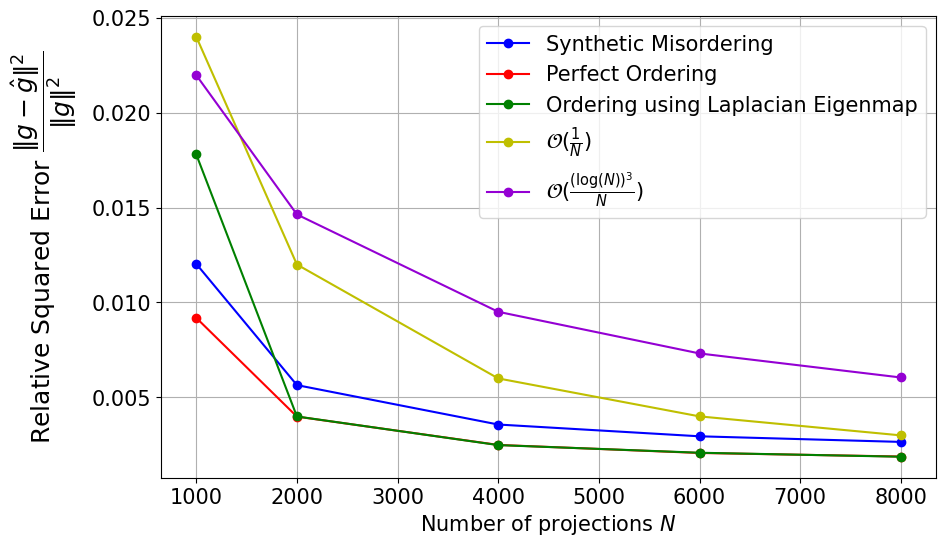}
     \includegraphics[width=0.48\textwidth]{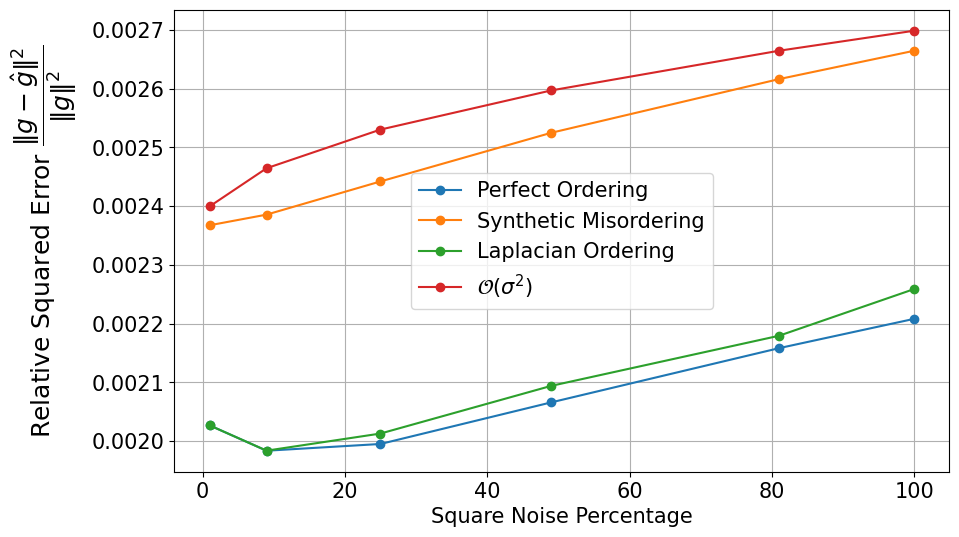}
     \includegraphics[width=0.48\textwidth]{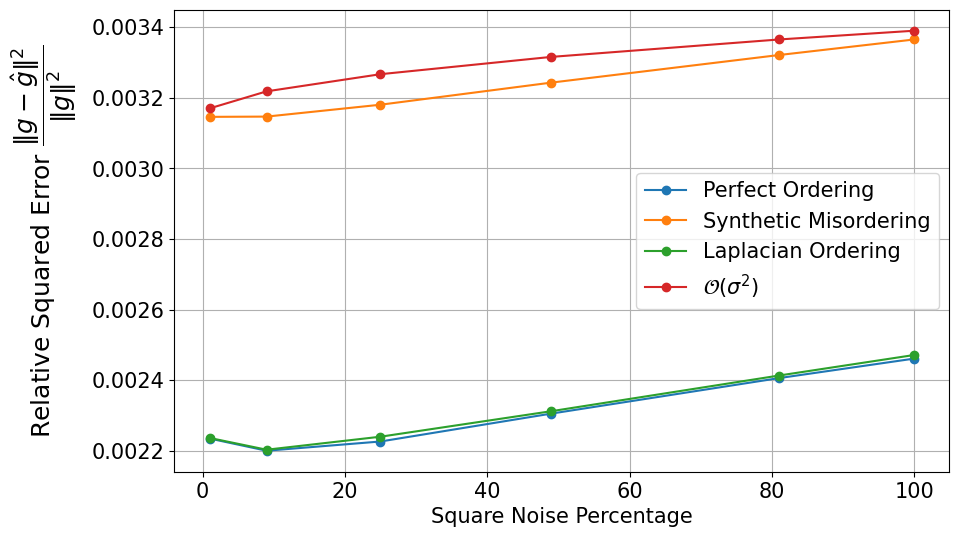} 
     \includegraphics[width=0.48\textwidth]{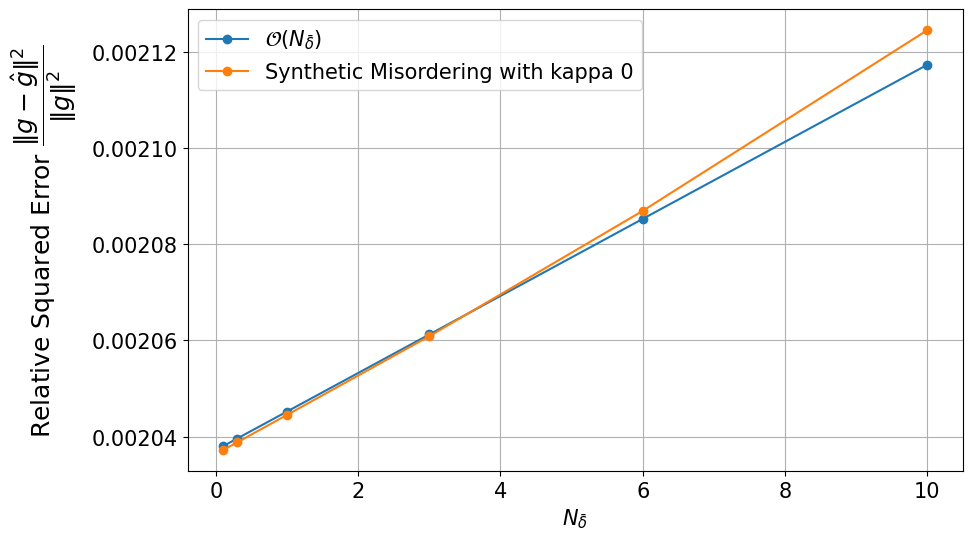}   \includegraphics[width=0.48\textwidth]{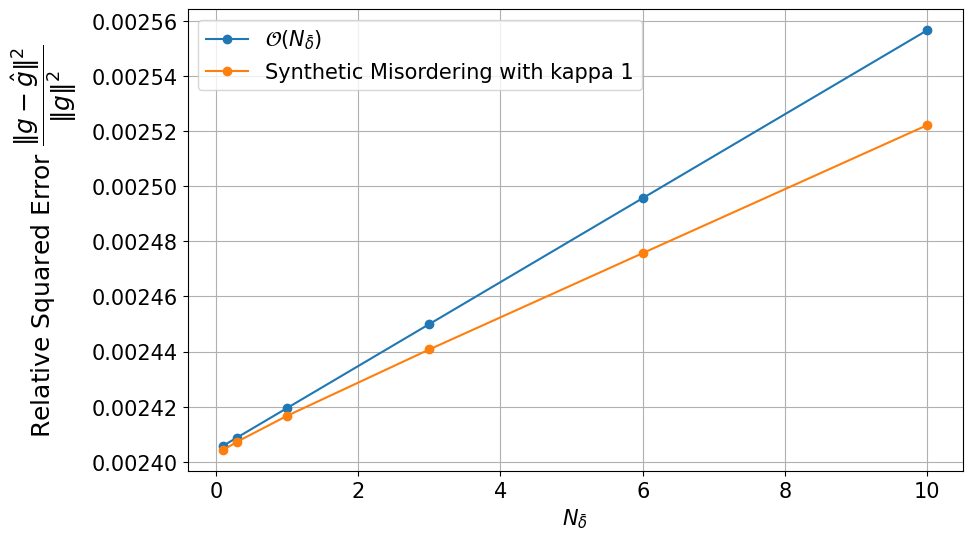}     
     \caption{In the first two rows, the plots are presented for settings S1, S2, S3 (defined in the text) for angles drawn from $\textrm{Uniform}(0,2\pi)$ (left) and $\textrm{VM}(\mu=\pi,\kappa=1)$ (right). First row: Squared relative reconstruction error $E$ (averaged over 10 runs) versus the number of projections $N$ keeping $f_n = 0.05$. A plot of $\bigo{(1/N)}$ and $\mathcal{O}((\log N)^3/N)$ are shown for comparison. See \texttt{SM} supplemental material for plots for $\kappa = 2$. Second row: Plot of $E$ versus squared noise percentage $f_n ^ 2$ for $N = 5000$. Third row: Plot of $E$ versus $N_{\bar{\delta}}$ for $f_n = 0.05$ and $N = 5000$ for setting S2.}
    \label{fig:recon_errors1}
\end{figure}

\noindent\textbf{$E$ versus $N$:} Keeping $\sigma$ constant, the effect of increasing the number of projections $N$, on the quality of reconstruction obtained from variants of the algorithm \textsf{RA}, was verified under the following three settings: (S1) Noisy projections and perfect ordering information ($f_n = 0.05$), (S2) Noisy projections ($f_n = 0.05$) and \emph{imperfect} ordering information, generated synthetically via shuffles and shifts, i.e., a mapping $h \sim \textrm{good}(\bar{\delta},N_{\bar{\delta}})$ was synthetically generated with $\bar{\delta} :=\bigob{\sqrt{N}}, N_{\bar{\delta}}:=\bigob{(\log N)^2}$, (S3) Noisy projections ($f_n = 0.05$) and (possibly imperfect) ordering information as obtained by the LE algorithm. PCA based denoising was employed in all scenarios involving noisy projections. In all settings, the value of $\nu_0$ was chosen so as to minimize the cross validation error on a held out set of projections. The squared relative reconstruction error was computed using the formula $E := \|\hat{g}-g\|^2/\|g\|^2$, where $g$ is the true image and $\hat{g}$ is its estimate. These error values are plotted in Fig.~\ref{fig:recon_errors1} (top row), where a clear decreasing trend is observed, with experimental errors upper bounded by $\bigo{(\log^3 N/N)}$ in all cases for $\kappa = 0$ and $\kappa = 1$. This is in tune with Thm.~\ref{thm:2d_reconstruction} for a fixed $\sigma$. 

\begin{figure*}[htb]
    \centering
    \includegraphics[width=0.1\textwidth]{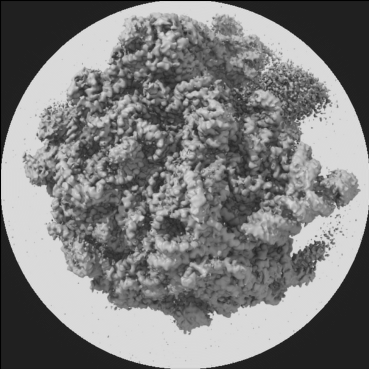}
    \includegraphics[width=0.44\linewidth]{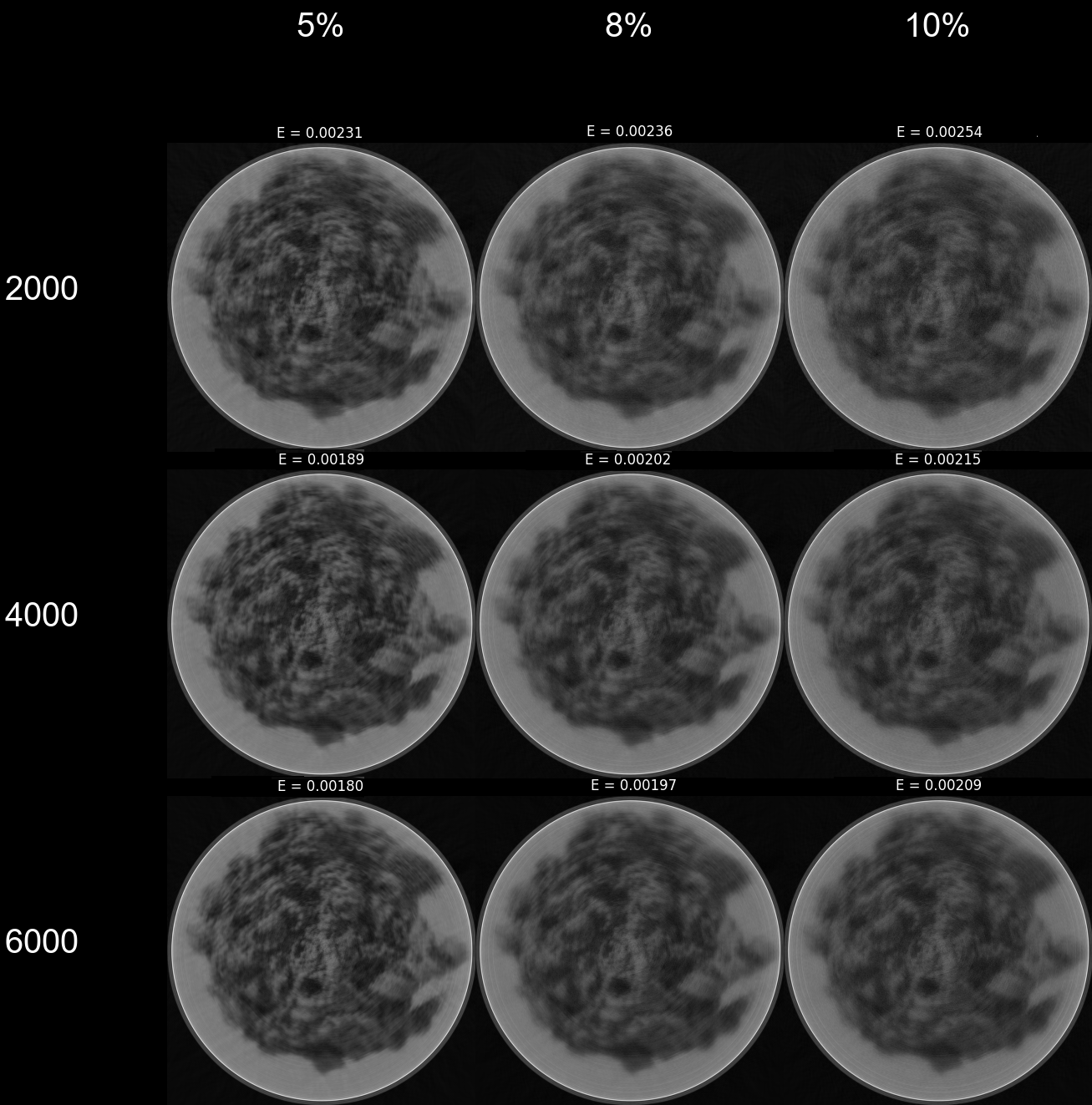}
    \includegraphics[width=0.44\linewidth]{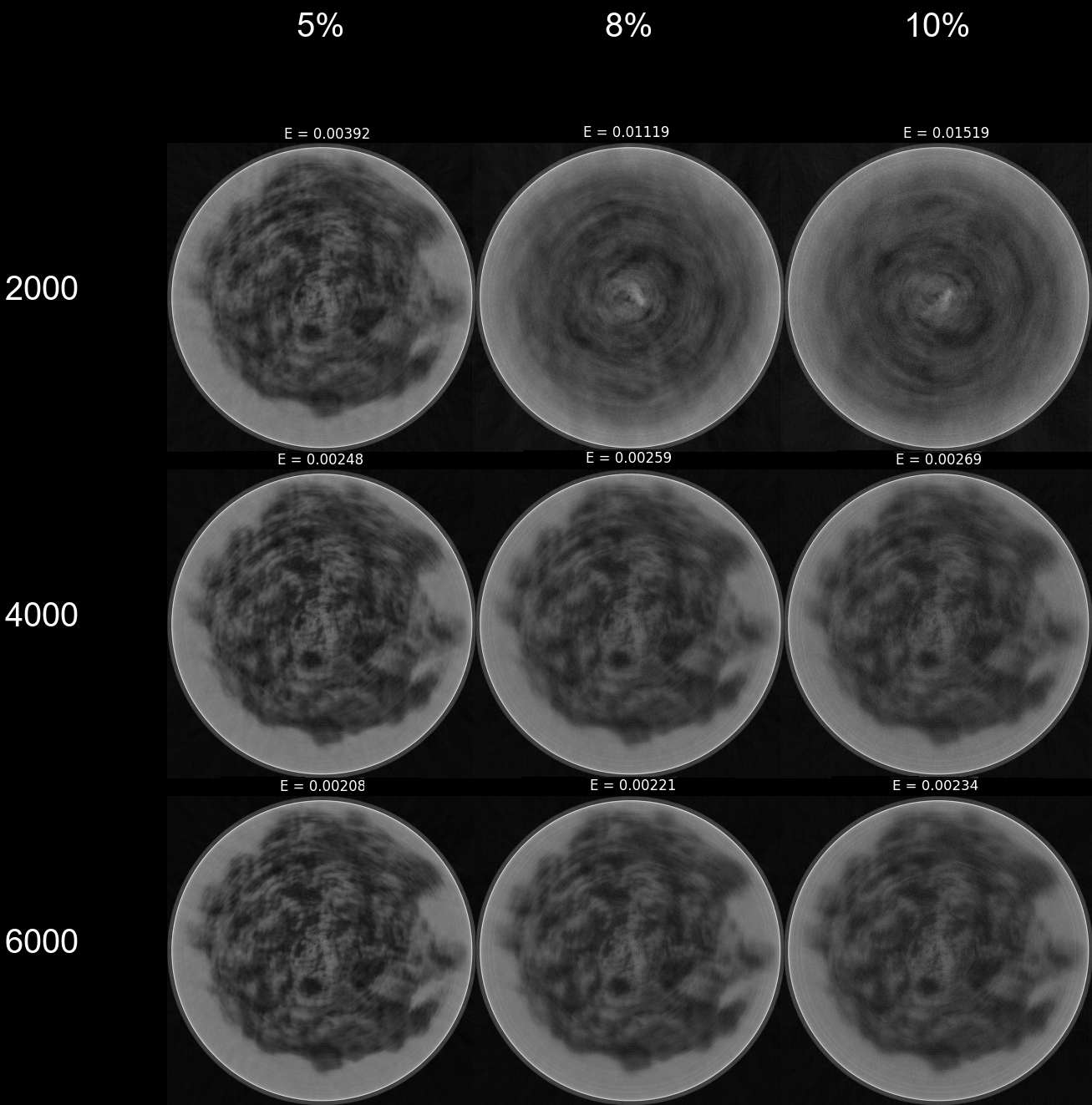}
   \caption{Leftmost: ground truth image; Sample reconstructions and relative reconstruction error ($E$) values with $N \in \{2000,4000,6000\}$ projections and $f_n \in \{0.05,0.08,0.1\}$ for setting S3, i.e. via Laplacian eigenmaps, for angles drawn from $\textrm{Uniform}(0,2\pi)$ (left) and $\textrm{VM}(\pi,1)$ (right). Results with other $\kappa$ values are presented in supplemental material.}
\label{fig:recons_noise0}
\end{figure*}

\noindent\textbf{$E$ versus $\sigma^2$:} Likewise, plots of $E$ versus $f^2_n$ (i.e. effectively versus $\sigma^2$) for a fixed $N = 5000$ are also presented in Fig.~\ref{fig:recon_errors1} (second row). These plots demonstrated that $E$ is upper bounded by $\bigo{(\sigma^2)}$ as shown in Thm.~\ref{thm:2d_reconstruction} for a constant $N = 5000$. The plots also show that for $f_n \leq 0.1$, the LE technique with a prior denoising step (see description of \textsf{RA} at the end of Sec.~\ref{sec:tomog}) produces near-perfect ordering. Sample image reconstruction results along with values for $E$ are shown in Fig.~\ref{fig:recons_noise0} for $N \in \{2000,4000,6000\}$ and $f_n \in \{0.05, 0.08, 0.1\}$. Here the reconstruction quality is seen to improve with increase in $N$ and worsen with increase in $f_n$. 

\noindent\textbf{$E$ versus $N_{\bar{\delta}}$:} Lastly, the linear increase in $E$ with increase in $N_{\bar{\delta}}$ for $\bar{\delta} = 4$ is shown in Fig.~\ref{fig:recon_errors1} (third row) for $N = 5000, f_n = 0.05$. This again tallies with Thm.~\ref{thm:2d_reconstruction}. 

\begin{figure}[htb]
    \centering
    \includegraphics[width=0.4\textwidth]{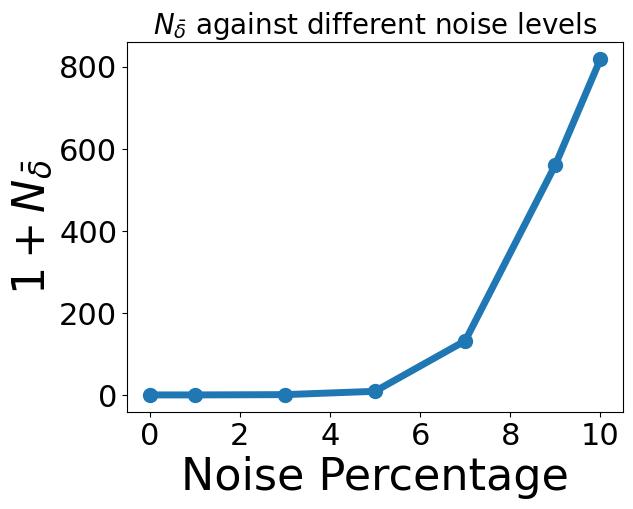}
    \includegraphics[width=0.4\textwidth]{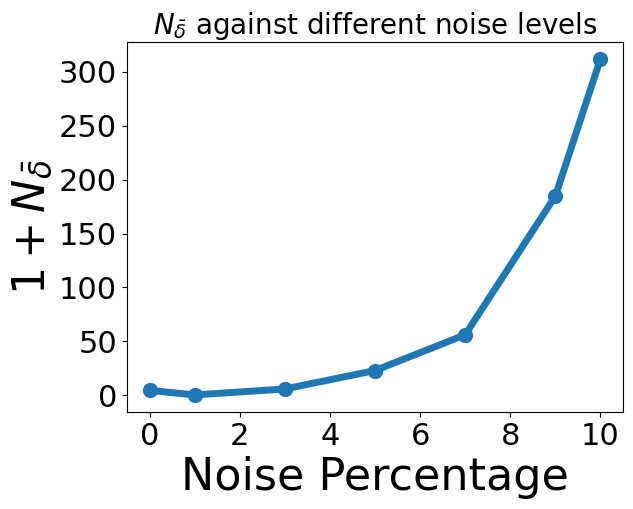}
    \includegraphics[width=0.4\textwidth]{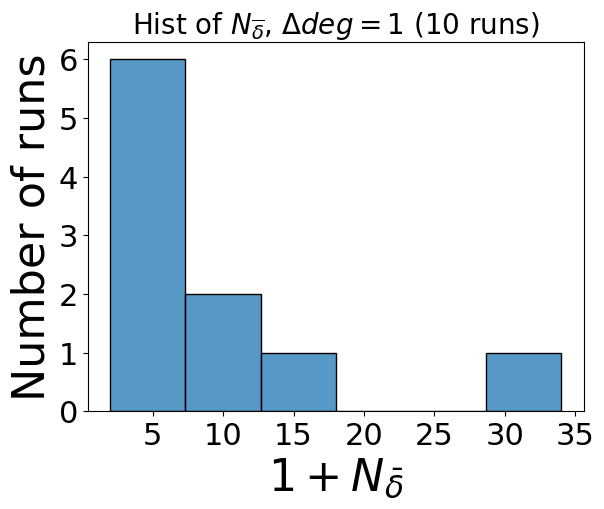}
    \includegraphics[width=0.4\textwidth]{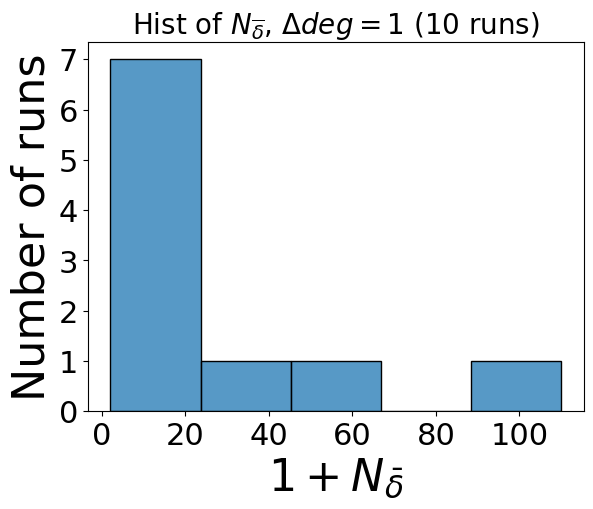}
    \caption{Variation of $N_{\bar{\delta}}$ averaged over 10  instances with noise percentage $f_n$ (first row). Histograms of $N_{\bar{\delta}}$ values over the different instances for $f_n = 0.05$ (second row). In both cases, the angles are drawn from $\textrm{Uniform}(0,2\pi)$ (left) and $\textrm{VM}(\pi,1)$ (right). All results are for $N = 5000$ projections.}
    \label{fig:Ndelta_sigma}
\end{figure}

\noindent\textbf{Effect of $\sigma, F(.)$ on $N_{\bar{\delta}}$:} It is natural to expect that $N_{\bar{\delta}}$ would increase with $\sigma$. The variation of $N_{\bar{\delta}}$ with $\sigma$ is shown in Fig.~\ref{fig:Ndelta_sigma} -- in terms of a plot of average $N_{\bar{\delta}}$ values across 10 instances w.r.t. $\sigma$ and histograms of $N_{\bar{\delta}}$ values across the same noise instances. These plots are for angle distributions given by $\textrm{Uniform}(0,2\pi)$ and $\textrm{VM}(\pi,1)$, and are for $N = 5000$ and $\bar{\delta}$ corresponding to a 1 degree difference. We see that $N_{\bar{\delta}}$ remains constant for a range of $\sigma$ values, and increases thereafter, but remains very small compared to $N$. Moreover, the values of $N_{\bar{\delta}}$ are smaller for $\textrm{Uniform}(0,2\pi)$ as compared to $\textrm{VM}(\pi,1)$, (though both are significantly smaller than $N$) which is to be expected as the latter distribution allows fewer samples from certain regions of the unit circle. In general, we have empirically observed smaller $N_{\bar{\delta}}$ values for distributions that are closer to uniform.

\section{Conclusion and Directions for Future Work}
\label{sec:conclusion}
We have presented a theoretical analysis of reconstruction errors for 2D tomography given 1D projections at unknown angles drawn from a known distribution. Our analysis builds on recent results on band-limited signal reconstruction given unknown sampling locations but from a known distribution. Though the associated reconstruction algorithms for 2D UVT are well known \cite{Coifman2008}, ours is the first approach to present asymptotic theoretical bounds for this problem without restrictive assumptions on the structure of the underlying image. Our work only assumes that the image is QBL and that it obeys a certain kind of power law (commonly obeyed by images that lie in a Sobolev space, but with a few other technical restrictions \cite{Beckmann2015}).

Our analysis currently requires knowledge of the angle distribution. It is indeed possible to estimate the distribution on the fly from the projections while performing image reconstruction, but results on this will be reported elsewhere. At this point, we have empirically verified that the ordering errors produced by the LE-based ordering algorithm (with a suitable pre-processing step for denoising) are $o(N)$ for a range of noise levels and distributions, but have not theoretically proved it. We have experimented with noise levels upto $f_n = 0.1$. For higher noise levels, the LE-based ordering will produce errors, but we believe that these errors can be controlled with better denoising algorithms, such as those based on machine learning approaches. We did not explore this direction, as the main emphasis of our paper was on theoretical bounds for the 2D UVT problem. 

Furthermore, extending the theoretical analysis to 3D is non-trivial. In 3D, each projection image would be associated with a 3D rotation matrix, or equivalently with an axis of rotation and an angle of rotation in the plane perpendicular to the axis. Both the axis and the angle can be represented as points on a sphere. This would require arrangement and sorting of projections along longitudinal lines on a sphere and sorting of the longitudinal lines. This is an important direction for future work. 


\newpage
\customtitle{Supplemental Material}
\appendix
\begin{corollary}
\label{cor:disc_recon}
For $N > e^{2\pi \nu_0 r_0}$ and $h \sim \textrm{good}(\bar{\delta}, N_{\bar{\delta}})$,
\begin{align}
 \int_0^{\nu_0} \xpcttnb{\left\| \mathcal{F}R^{\theta}_g(\nu) - \widehat{\mathcal{F}R}^{\theta}_g(\nu)\right\|_2^2} d\nu = \bigob{\frac{\nu_0 \sigma^2 k_0}{N\zeta^2}} + \bigob{\frac{\nu_0 N_{\bar{\delta}}k_0}{N \zeta^2}} \nonumber \\+ \bigob{\frac{\nu_0 k_0}{N^2 \zeta^2}} + \bigob{\frac{\nu_0 k^3_0}{N \zeta^4} + \frac{\nu_0 k^3_0 \bar{\delta}^2}{N^2 \zeta^4}} =\bigob{\frac{\nu_0(\sigma^2 + N_{\bar{\delta}})\log N}{N \zeta^2} + \frac{\nu_0 \log^3 N}{N \zeta^4} + \frac{\nu_0 \log^3 N \bar{\delta}^2}{N \zeta^4}}.       
\end{align}
\end{corollary}

Note that reconstructing a ring of $\mathcal{F}R^{\theta}_g(\nu_1)$ is equivalent to reconstructing a ring of $\mathcal{F}T_g$ with radius $\nu_1$ (as discussed in the main paper), and hence if we perform the reconstruction in Lemma~\ref{lem:fr_ring_reconstruction} for every $\nu_1 \leq \nu_0$ (where $\nu_0$ is an arbitrarily chosen radius), we can reconstruct the disc of radius $\nu_0$ in $\mathcal{F}T_g$.

Thus our reconstruction can be defined as follows:
\begin{align}
     \widehat{\mathcal{F}T}_g(u,v) &:= \begin{cases}
        \widehat{\mathcal{F}R}^{\theta}_g(\nu=\sqrt{u^2+v^2}) & u^2+v^2 \leq \nu_0^2, \theta=\arctan(\frac{v}{u})\\
        0 & \textrm{otherwise}
     \end{cases}\\
     \hat{g} &:= \mathcal{F}^{-1}(\widehat{\mathcal{F}T}_g),
     \label{eqn_frecon_defn}
\end{align}
where $\mathcal{F}^{-1}$ stands for inverse Fourier Transform. Since we are considering bandlimited approximation of a QBL signal, we now present a result to bound the tail of the Fourier transform magnitude outside the radius $\nu_0$, for an image which obeys a certain kind of power law as described in \cite{Beckmann2015}. 

\begin{lemma}
    \label{lem:sobolev}
    Consider $\alpha \in \mathbb{R}_{+} \cup \{0\}$ and $g:\Omega \subset \mathbb{R}^2 \rightarrow \mathbb{R}$. Define $\|g\|_\alpha^2 := \frac{1}{2\pi} \int\int (1+u^2+v^2)^\alpha |\mathcal{F}T_g(u, v)|^2 du dv$. If $\|g\|_\alpha^2 < \infty$,
    then for any $\nu_0>0$ the tail of the magnitude of $\mathcal{F}T_g$ outside the radius $\nu_0$ is bounded by:
    \begin{equation}
       e_g(\nu_0) := \frac{1}{2\pi} \int_{u^2+v^2>\nu_0^2} |\mathcal{F}T_g(u, v)|^2 dv du \leq \nu_0^{-2\alpha} \|g\|_\alpha^2.
    \end{equation}
\end{lemma}

\begin{theorem}
    Consider some arbitrarily chosen radius $\nu_0$. For $N > e^{2\pi \nu_0 r_0}$, for $k_0 := \left\lceil \frac{\log N}{0.765} \right\rceil$, $g$ which satisfies the conditions in Lemma~\ref{lem:sobolev}, and the reconstruction $\hat{g}$ defined in Eqn.~\ref{eqn_frecon_defn},
    \begin{align}
        \xpcttn \left[ \| g - \hat{g} \|^2 \right] &= \bigob{\frac{\nu_0 \sigma^2 k_0}{N\zeta^2}} + \bigob{\frac{\nu_0 N_{\bar{\delta}}k_0}{N \zeta^2}}  + \bigob{\frac{\nu_0 k_0}{N^2 \zeta^2}} + \bigob{\frac{\nu_0 k^3_0}{N \zeta^4} + \frac{\nu_0 k^3_0 \bar{\delta}^2}{N^2 \zeta^4}} +  \bigob{\nu_0^{-2\alpha}\|g\|_\alpha^2}\nonumber \\
        &= \bigob{\frac{\nu_0(\sigma^2 + N_{\bar{\delta}})\log N}{N \zeta^2} + \frac{\nu_0 \log^3 N}{N \zeta^4} + \frac{\nu_0 \log^3 N \bar{\delta}^2}{N^2 \zeta^4}} + \bigob{\nu_0^{-2\alpha}\|g\|_\alpha^2}.
    \end{align}
    \label{thm:2d_reconstruction}
\end{theorem}

\section{Proof for Theorem ~\ref{thm:1d_reconstruction} for 1D Signal Reconstruction}
Consider a band-limited approximation of the original quasi-bandlimited signal $g \sim \textrm{qbl}(k_1, \gamma)$ with $k_0 := \lceil \frac{\log(N)}{\gamma} \rceil$ as the bandwidth of this approximation namely,
\begin{align}
    g^{k_0}(t) &:= \sum_{|k| \leq k_0} a_k e^{j2\pi kt}.
\end{align}
Referring to the notation used in the definition of \textsf{(P3)}, our reconstruction from Thm.~\ref{thm:1d_reconstruction} is as follows:
\begin{align}
    \Bar{a}_k &:= \sum_{i=1}^N (g(t'_i)+\varepsilon(t'_i))(x_i - x_{i-1}) \exp \left( -j2\pi k x_i \right),\\
    \Bar{g}^{k_0}(t) &:= \sum_{k=-k_0}^{k_0} \Bar{a}_k e^{j2\pi kt},
\end{align}
where $x_i := F^{-1}\left( \frac{i}{N} \right), x_0 := 0$ where $f(.), F(.)$ stand for the PDF and CDF of the (unknown) sampling locations. We make the assumption that $F(.)$ is invertible everywhere, and that $f(.)$ is lower bounded by a value $\zeta > 0$. 

As intermediate steps in our analysis, we also use
\begin{align}
    \label{eqn_atilde_defn}
    \Tilde{a}_k &:= \sum_{i=1}^N g(t'_i)(x_i - x_{i-1}) \exp \left( -j2\pi k x_i \right)\\
    \label{eqn_gtilde_defn}
    \Tilde{g}^{k_0}(t) &:= \sum_{k=-k_0}^{k_0} \Tilde{a}_k e^{j2\pi kt}\\
    \label{eqn_ahat_defn}
    \hat{a}_k &:= \sum_{i=1}^N g(t_i)(x_i - x_{i-1}) \exp \left( -j2\pi k x_i \right)\\
    \label{eqn_ghat_defn}
    \hat{g}^{k_0}(t) &:= \sum_{k=-k_0}^{k_0} \hat{a}_k e^{j2\pi kt}.
\end{align}
In particular note that $g(t)$ is the true signal, $g^{k_0}(t)$ is its true bandlimited approximation, $\bar{g}^{k_0}(t)$ is the bandlimited reconstruction from noisy samples with estimated ordering, $\tilde{g}^{k_0}(t)$ is the bandlimited reconstruction from noiseless samples with estimated ordering, and $\hat{g}^{k_0}(t)$ is the bandlimited reconstruction from noiseless samples with correct (ground truth) ordering assuming that the sample $g(t_i)$ is obtained at location $i/N$. Also, $a_k, \bar{a}_k, \tilde{a}_k, \hat{a}_k$ are the Fourier series coefficients corresponding to $g(t), \bar{g}^{k_0}(t), \tilde{g}^{k_0}(t), \hat{g}^{k_0}(t)$ respectively.

We first prove the following preliminary results.

\begin{lemma}\label{lem:overall_breakdown}
    \begin{align*}
        \xpcttnb{\|g(t) - \Bar{g}^{k_0}(t)\|_2^2} &\leq 4 \xpcttnb{\|g(t) - g^{k_0}(t)\|_2^2} \textup{(Error from bandlimited approximation)}\\
        &+ 4 \xpcttnb{\|g^{k_0}(t) - \hat{g}^{k_0}(t)\|_2^2} \textup{(Error from unknown location)}\\
        &+ 4 \xpcttnb{\|\hat{g}^{k_0}(t) - \Tilde{g}^{k_0}(t)\|_2^2} \textup{(Error from misspecified ordering)}\\
        &+ 4 \xpcttnb{\|\Tilde{g}^{k_0}(t) - \Bar{g}^{k_0}(t)\|_2^2}. \textup{(Error from noisy recovery)} \numberthis \label{eqn:overall_breakdown}
    \end{align*}
\end{lemma}
\begin{proof}
    Consider $\|g(t) - \Bar{g}^{k_0}(t)\|_2^2 = \|(g(t)-g^{k_0}(t))+(g^{k_0}(t)-\hat{g}^{k_0}(t)) + (\hat{g}^{k_0}(t)-\tilde{g}^{k_0}(t)) + (\Tilde{g}^{k_0}(t) - \Bar{g}^{k_0}(t))\|^2_2$. The lemma then directly follows from the  property of $\ell_2$ norm, where given a set $\{\boldsymbol{x}_i(t)\}_{i=1}^n$ of signals, we have $\left\| \sum_{i=1}^n \boldsymbol{x_i} (t)\right\|_2^2 \leq n \sum_{i=1}^n \left\| \boldsymbol{x_i}(t) \right\|_2^2$.
\end{proof}

\begin{lemma}\label{lem:g_gk0_close}
For $N \geq e^{\gamma k_1}$ we have: 
\begin{equation}
     \|g(t) - g^{k_0}(t)\|_2^2 = \bigo \left( \frac{1}{N^2} \right) \label{eqn_g_gk0_close},
\end{equation}
and $\forall t$:
\begin{equation}
    \label{eqn_g_gk0_close_at_t}
    |g(t) - g^{k_0}(t)| = \bigo \left( \frac{1}{N} \right). 
\end{equation}
\end{lemma}
\begin{proof}
Given the definition $k_0 = \left\lceil \frac{\log(N)}{\gamma} \right\rceil$, it follows that, $N \geq e^{\gamma k_1} \implies k_0 \geq k_1$. Using the QBL property of $g(t)$ and bounding the coefficients $a_k$ for $ |k| > k_0$, we get:
\begin{align}
    \|g(t) - g^{k_0}(t)\|_2^2 &\mycomp{(a)}{=} \sum_{|k| > k_0} |a_k|^2 \mycomp{(b)}{\leq} \sum_{|k| > k_0} (de^{-\gamma|k|})^2\\
    &= 2d^2 \sum_{k > k_0} e^{-2\gamma k} \mycomp{(c)}{=} 2d^2 \cdot \dfrac{e^{-2\gamma k_0}}{1 - e^{-2}}\\
    &= \bigo \left( e^{-2\gamma k_0} \right) \mycomp{(d)}{=} \bigo \left( e^{-\log(N^2)} \right)\\
    &= \bigo \left( \frac{1}{N^2} \right),
\end{align}
where (a) directly follows by using Parseval's theorem on $g(t) - g^{k_0}(t) = \sum_{|k| > k_0} a_k e^{j2\pi kt}$, (b) uses the definition of $g \sim \textrm{qbl}(k_1, \gamma)$ for $|k| > k_0 \geq k_1$, (c) is a result of using the formula for the sum of a geometric progression, and (d) uses the definition $k_0 = \log N/\gamma$. Notice that this bounds the first term of Lemma~\ref{lem:overall_breakdown}.\\
Similarly,
\begin{align}
    \forall t: \left|g(t) - g^{k_0}(t)\right| &= \left| \sum_{|k| > k_0} a_k e^{j2\pi k t} \right| \leq \sum_{|k| > k_0}  \left| a_k e^{j2\pi k t} \right|\\
    &= \sum_{|k| > k_0}  \left| a_k \right| \leq \sum_{|k| > k_0} (de^{-\gamma|k|})\\
    &= 2d \sum_{k > k_0} e^{-\gamma k} \mycomp{(a)}{=} 2d \cdot \dfrac{e^{-\gamma k_0}}{1 - e^{-1}}\\
    &= \bigo \left( e^{- \gamma k_0} \right) = \bigo \left( e^{-\log N} \right)\\
    &= \bigo \left( \frac{1}{N} \right).
\end{align}
\end{proof}

\begin{lemma}
    \begin{align}
        \label{eqn_g_k0_smooth}
        \left| \frac{dg^{k_0}(t)}{dt} \right| \leq 2\pi k_0,\\
        \label{eqn_ordr_stats_close2}
        N\xpcttnb{\left|t_i - x_i \right|^2} &\leq \frac{1}{4 f^2(x_i)} + \bigo(\sqrt{1/N}),\\
        \begin{split} \label{eqn_g_k0_difference}
            \xpcttnb{\left|g^{k_0}(t_i)-g^{k_0}\left(x_i\right)\right|^2} &\leq \left\|\frac{dg^{k_0}(t)}{dt}\right\|_\infty^2 \xpcttnb{\left|t_i - x_i\right|^2}\\
            &\leq (2\pi k_0)^2 \left( \frac{1}{4f^2(x_i)N} + \bigo \left( \frac{1}{N\sqrt{N}} \right) \right)\\
            &= \bigob{\frac{k_0^2}{f^2(x_i) N}} \leq \bigob{\dfrac{k^2_0}{\zeta^2 N}}.
        \end{split}
    \end{align}
\end{lemma}
\begin{proof}
    Eqn.~(\ref{eqn_g_k0_smooth}) follows from Eqn.~(\ref{eqn_gradient_bound}) for $g^{k_0}(t)$ which has bandwidth $k_0$. We consider the well known result (see for example \cite[Sec. 1]{Walker1968}) that $t_i - x_i \sim \mathcal{N}(0,p_i(1-p_i)/(Nf^2(x_i)))$ for large $N$. Here $p_i := F(x_i) = i/N$. Hence $NE[(t_i-x_i)^2] = \dfrac{p_i(1-p_i)}{f^2(x_i)} + O(1/\sqrt{N}) \leq \dfrac{1}{4f^2(x_i)} + O(1/\sqrt{N})$.   
    Eqn.~(\ref{eqn_ordr_stats_close2}) is the equivalent of Eqn.~(9) from the main paper for non-uniform distributions.     
    Eqn.~(\ref{eqn_g_k0_difference}) follows from using Lagrange's mean value theorem with Eqn.~(\ref{eqn_g_k0_smooth}) and Eqn.~(\ref{eqn_ordr_stats_close2}). Note that $t_i$ are not known during reconstruction, but these results help us bound $\xpcttnb{\|g^{k_0}(t) - \hat{g}^{k_0}(t)\|_2^2}$, which is an intermediate step in our analysis (as indicated by Lemma~\ref{lem:overall_breakdown}). The final inequality follows from the assumption that $f(\cdot) \geq \zeta > 0$.
\end{proof}

\begin{lemma}
    \label{lem:parseval_breakdown}
    \begin{align}
        \xpcttnb{\|g^{k_0}(t) - \hat{g}^{k_0}(t)\|_2^2} &= \sum_{k=-k_0}^{k_0} \xpcttnb{|a_k - \hat{a}_k|^2}\\
        \xpcttnb{\|\hat{g}^{k_0}(t) - \Tilde{g}^{k_0}(t)\|_2^2} &= \sum_{k=-k_0}^{k_0} \xpcttnb{|\hat{a}_k - \Tilde{a}_k|^2}\\
        \xpcttnb{\|\Tilde{g}^{k_0}(t) - \Bar{g}^{k_0}(t)\|_2^2} &= \sum_{k=-k_0}^{k_0} \xpcttnb{|\Tilde{a}_k - \Bar{a}_k|^2}
    \end{align}
\end{lemma}
\begin{proof}
    These directly follow from Parseval's theorem and the definitions of $g^{k_0}(t), \hat{g}^{k_0}(t), \Tilde{g}^{k_0}(t), \Bar{g}^{k_0}(t)$.
\end{proof}

\begin{lemma}
    \label{lem:noise_term_bound}
    \begin{equation}
        \xpcttnb{\|\Tilde{g}^{k_0}(t) - \Bar{g}^{k_0}(t)\|_2^2} = \bigob{\frac{k_0 \sigma^2}{N\zeta^2}}
    \end{equation}
\end{lemma}
\begin{proof}
    From the definitions of $\Tilde{a}_k$ and $\Bar{a}_k$, we have
    \begin{align*}
        \Bar{a}_k &= \Tilde{a}_k + \sum_{i=1}^N \varepsilon(t'_i) (x_i - x_{i-1}) \exp \left( -j2\pi k x_i \right).\\
        \therefore \xpcttnb{|\Tilde{a}_k - \Bar{a}_k|^2} &= \xpcttnb{\left| \sum_{i=1}^N \varepsilon(t'_i) (x_i - x_{i-1}) \exp \left( -j2\pi k x_i \right) \right|^2}\\
        &\mycomp{(a)}{=}\sum_{i=1}^N \xpcttnb{\left|\varepsilon(t'_i) (x_i - x_{i-1}) \exp \left( -j2\pi k x_i \right) \right|^2}\\
        &=\sum_{i=1}^N \xpcttnb{\left|\varepsilon(t'_i) (x_i - x_{i-1})\right|^2}\\
        &= \sum_{i=1}^N \xpcttnb{\left|\varepsilon(t'_i)\right|^2} \left|(x_i - x_{i-1})\right|^2 = \sigma^2 \sum_{i=1}^N (x_i - x_{i-1})^2\\
        &\leq \dfrac{\sigma^2}{N \zeta^2},
    \end{align*}
    where (a) uses the fact that as $\{\varepsilon(t'_i)\}_{i=1}^N$ are independent of each other, and moreover $\{(x_i - x_{i-1})^2\}_{i=1}^N$ are constants. Due to this, the variance of their sum is the sum of their individual variances. The last inequality uses 
    Lemma~\ref{lem:gap_order_stats} stated and proved later. Applying Lemma~\ref{lem:parseval_breakdown} then completes the proof (by multiplying each term on the RHS of the present expression by $k_0$). Notice that this bounds the fourth term of Lemma~\ref{lem:overall_breakdown}. 
\end{proof}

We now seek to bound the third term of Lemma~\ref{lem:overall_breakdown}. Recall that $t'_i = t_{h(i)}$ where $h \sim \textrm{good}(\bar{\delta}, N_{\bar{\delta}})$ is the bijective mapping defined in the main paper. Define
\begin{align}
    \mathcal{S}_1 &:= \{ i \in [N] : |i - h(i)| > \bar{\delta} \},\\
    \mathcal{S}_2 &:= [N] - \mathcal{S}_1 = \{ i \in [N] : |i - h(i)| \leq \bar{\delta} \}.
\end{align}
We bound $\xpcttnb{\|\hat{g}^{k_0}(t) - \Tilde{g}^{k_0}(t)\|_2^2}$ using the following results.
\begin{lemma}
    \label{lem:ahat_aprime_prelim}
    \begin{align}
        \label{eqn_ahat_aprime_prelim}
        \xpcttnb{|\hat{a}_k - \Tilde{a}_k|^2} &= \bigob{\frac{N_{\bar{\delta}}}{N\zeta^2}} + \frac{1}{N\zeta^2} \sum_{\mathcal{S}_2} \xpcttnb{\left|g(t'_i) - g(t_i) \right|^2}.
    \end{align}
\end{lemma}

\begin{proof}
    \begin{align*}
        |\hat{a}_k - \Tilde{a}_k|^2 &= \left| \sum_{i=1}^N (g(t^{\prime}_i) - g(t_i)) (x_i - x_{i-1}) \exp \left( -j2\pi k x_i \right) \right|^2\\
        &\mycomp{(a)}{\leq} N \sum_{i=1}^N \left|(g(t^{\prime}_i) - g(t_i))  (x_i - x_{i-1}) \exp \left( -j2\pi k x_i \right) \right|^2\\
        &= N\sum_{i=1}^N \left|g(t^{\prime}_i) - g(t_i) \right|^2 \left|x_i - x_{i-1} \right|^2 \\
        &= N\sum_{\mathcal{S}_1} \left|g(t^{\prime}_i) - g(t_i) \right|^2 \left|x_i - x_{i-1} \right|^2 + N\sum_{\mathcal{S}_2} \left|g(t^{\prime}_i) - g(t_i) \right|^2 \left|x_i - x_{i-1} \right|^2  \\
        &\mycomp{(b)}{\leq} N \sum_{\mathcal{S}_1} 2^2 \left|x_i - x_{i-1} \right|^2  + N \sum_{\mathcal{S}_2} \left|g(t^{\prime}_i) - g(t_i) \right|^2 \left|x_i - x_{i-1} \right|^2 \\
        &\mycomp{(c)}{\leq} \frac{4N_{\bar{\delta}}}{N\zeta^2} + \frac{1}{N \zeta^2} \cdot \sum_{\mathcal{S}_2} \left|g(t^{\prime}_i) - g(t_i) \right|^2
    \end{align*}
    where (a) follows from the inequality $\left| \sum_{i=1}^N x_i \right|^2 \leq N \sum_{i=1}^N |x_i|^2$, (b) follows because $|g(t)| < 1$ and (c) follows because $h \sim \textrm{good}(\bar{\delta}, N_{\bar{\delta}}) \implies |\mathcal{S}_1| \leq N_{\bar{\delta}}$, and $|x_i - x_{i-1}| \leq \frac{1}{N\zeta}$ from Lemma~\ref{lem:gap_order_stats}. Taking expectation on both sides of the expression completes the proof.
\end{proof}

\begin{lemma}
For $\forall i$, we get:
    \label{lem:g_misorder_to_gk0_misorder}
    \begin{equation}
        \begin{split}
            \xpcttn \left[ |g(t_{h(i)}) - g(t_i)|^2 \right] &= \bigob{\frac{1}{N^2}} + 3 \xpcttnb{|g^{k_0}(t_{h(i)}) - g^{k_0}(t_i)|^2}.
        \end{split}
    \end{equation}
    
\end{lemma}
\begin{proof}
Consider $\left|g(t_{h(i)}) - g(t_i)\right|^2 = \left|g(t_{h(i)})-g^{k_0}(t_{h(i)})+g^{k_0}(t_{h(i)}) - g^{k_0}(t_i) + g^{k_0}(t_i) - g(t_i)\right|^2$. Using the inequality $\left| \sum_{i=1}^N x_i \right|^2 \leq N \sum_{i=1}^N |x_i|^2$, we find:
    \begin{align*}
        &\xpcttnb{|g(t_{h(i)}) - g(t_i)|^2} \leq 3\big(\xpcttnb{ |g(t_{h(i)}) - g^{k_0}(t_{h(i)})|^2}\\
        &\qquad + \xpcttnb{ |g^{k_0}(t_{h(i)}) - g^{k_0}(t_i)|^2} + \xpcttnb{ |g^{k_0}(t_i) - g(t_i)|^2} \big)\\
        &\mycomp{(a)}{=} \bigob{\frac{1}{N^2}} + 3 \xpcttn \big[|g^{k_0}(t_{h(i)}) - g^{k_0}(t_i)|^2\big] + \bigob{\frac{1}{N^2}},
    \end{align*}
    where (a) follows from Eqn.~\ref{eqn_g_gk0_close_at_t} and Lemma~\ref{lem:g_gk0_close} applied to the first and third terms.
\end{proof}

\begin{lemma}
    \label{lem:g_ko_misorder_bound}
    For $i \in \mathcal{S}_2$, we have:
    \begin{align}
        \xpcttnb{|g^{k_0}(t_{h(i)}) - g^{k_0}(t_i)|^2} &= \bigob{\frac{k_0^2}{N \zeta^2}\left(1 + \frac{\bar{\delta}^2}{N}\right)}
    \end{align}
\end{lemma}

\begin{proof}
Observe that
    \begin{align*}
        &\xpcttnb{|g^{k_0}(t_{h(i)}) - g^{k_0}(t_i)|^2} = \xpcttn \Bigg[\bigg|g^{k_0}(t_{h(i)}) - g^{k_0}\left(x_{h(i)}\right)\\
        &\quad + g^{k_0}\left(x_{h(i)}\right) - g^{k_0}\left(x_i\right) + g^{k_0}\left(x_i\right) - g^{k_0}(t_i) \bigg|^2\Bigg]\\
        &\leq 3\xpcttn \Bigg[\bigg|g^{k_0}(t_{h(i)}) - g^{k_0}\left(x_{h(i)}\right)\bigg|^2 + \\
        &\quad \bigg| g^{k_0}\left(x_{h(i)}\right) - g^{k_0}\left(x_i\right)\bigg|^2 + \bigg|g^{k_0}\left(x_i\right) - g^{k_0}(t_i) \bigg|^2\Bigg]\\
        &\mycomp{(a)}{=} \bigob{\dfrac{k^2_0}{\zeta^2 N}} + 3\xpcttnb{\bigg| g^{k_0}\left(x_{h(i)}\right) - g^{k_0}\left(x_i\right)\bigg|^2} + \bigo \left( \dfrac{k^2_0}{\zeta^2 N} \right)\\
        &\mycomp{(b)}{\leq} \bigo \left( \dfrac{k^2_0}{\zeta^2 N} \right) + 3\left| (x_{h(i)} - x_i)  \left\|\frac{dg^{k_0}(t)}{dt}\right\|_\infty \right|^2\\
        &= \bigo \left( \dfrac{k^2_0}{\zeta^2 N} \right) + 3(2\pi k_0)^2 \left| x_{h(i)} - x_i \right|^2\\
        &\mycomp{(c)}{=} \bigo \left( \dfrac{k^2_0}{\zeta^2 N} \right) + \bigob{ \frac{k_0^2 \bar{\delta}^2}{N^2 \zeta^2} },
    \end{align*}
    where (a) follows from Eqn.~\ref{eqn_g_k0_difference} applied to the first and third terms, (b) from Lagrange's mean value theorem, and (c) follows from Lemma~\ref{lem:gap_order_stats} and because $h$ is a `good' mapping implying $|h(i)-i| \leq \bar{\delta}$.
\end{proof}

\begin{lemma}
    \label{lem:third_term_bound}
    \begin{align}
        \xpcttnb{\|\hat{g}^{k_0}(t) - \Tilde{g}^{k_0}(t)\|_2^2} = \bigob{\frac{N_{\bar{\delta}}k_0}{N\zeta^2} + 
        \frac{k_0}{N^2 \zeta^2} + 
        \frac{ k_0^3}{\zeta^4 N}\left(1+\frac{\bar{\delta}^2}{N}\right)}
    \end{align}
\end{lemma}

\begin{proof}
    Combining Lemmas~\ref{lem:ahat_aprime_prelim}, \ref{lem:g_misorder_to_gk0_misorder}, \ref{lem:g_ko_misorder_bound}, we get:
    \begin{align*}
        &\xpcttnb{|\hat{a}_k - \Tilde{a}_k|^2} = \bigob{\frac{N_{\bar{\delta}}}{N\zeta^2}} + \frac{1}{N\zeta^2} \sum_{\mathcal{S}_2} \left( \bigob{\frac{1}{N^2}} + \bigob{\frac{ k_0^2}{\zeta^2 N}\left(1+\frac{\bar{\delta}^2}{N}\right)} \right)\\
        &\qquad \mycomp{(a)}{=} \bigob{\frac{N_{\bar{\delta}}}{N\zeta^2} + 
        \frac{1}{N^2 \zeta^2} + 
        \frac{ k_0^2}{\zeta^4 N}\left(1+\frac{\bar{\delta}^2}{N}\right)}
    \end{align*}
    where (a) holds because $|\mathcal{S}_2| \leq N$. Applying Lemma~\ref{lem:parseval_breakdown} then completes the proof by multiplying each term on the RHS of the present expression by $k_0$ to obtain $\xpcttnb{\|\hat{g}^{k_0}(t)-\tilde{g}^{k_0}(t)\|^2}$. Notice that this bounds the third term of Lemma~\ref{lem:overall_breakdown}.
\end{proof}

We have bounded the first, third and fourth terms of Lemma~\ref{lem:overall_breakdown}. We still need to bound the second term. The proof for this proceeds along the lines of that in \cite{Kumar2015} but needs some modifications as $g(t)$ is no longer band-limited (cf. Lemmas~\ref{lem:quasi_breakdown2}, \ref{lem:quasi_breakdown1}).

\begin{lemma}
    \label{lem:quasi_breakdown}
    For all $|k| \leq k_0$, we have:
    \begin{equation}
        \label{eqn_quasi_breakdown}
        \begin{split}
             \left|\hat{a}_k - a_k\right|^2 \leq 
             2\left| \frac{1}{N} \sum_{i=1}^N \left(g(t_i)-g^{k_0}(x_i)\right)(x_i-x_{i-1)}) e^{-\frac{j2\pi ki}{N}}\right|^2
            + \nonumber \\
            2\left| \sum_{i=1}^N \left( g^{k_0}(x_i)(x_i-x_{i-1}) e^{ -\frac{j2\pi ki}{N} } - \int_{\frac{i-1}{N}}^{\frac{i}{N}} g^{k_0}(t) e^{-{j2\pi kt}} dt \right) \right|^2
        \end{split}
    \end{equation}
\end{lemma}
\begin{proof}
By definition of $\hat{a}_k$ and $a_k$, we have:
    \begin{align*}
        &\left|\hat{a}_k- a_k\right|\\
        &= \left| \sum_{i=1}^N g(t_i) (x_i - x_{i-1}) e^{-j2\pi k x_i} - \int_0^1 g^{k_0}(t) e^{-{j2\pi kt}} dt \right|\\
        &= \left| \sum_{i=1}^N g(t_i) (x_i - x_{i-1}) e^{-j2\pi k x_i} - \sum_{i=1}^N \int_{x_{i-1}}^{x_i} g^{k_0}(t) e^{-{j2\pi kt}} dt \right|\\
        &= \Bigg| \sum_{i=1}^N \left(g(t_i)-g^{k_0}\left( x_i \right)\right) (x_i - x_{i-1}) e^{-j2\pi k x_i} +\\
        &\quad \sum_{i=1}^N \left( g^{k_0}\left( x_i \right) (x_i - x_{i-1}) e^{ -j2\pi k x_i } - \int_{x_{i-1}}^{x_i} g^{k_0}(t) e^{-{j2\pi kt}} dt \right) \Bigg|\\
        &\leq \left| \sum_{i=1}^N \left(g(t_i)-g^{k_0}\left(x_i\right)\right) (x_i - x_{i-1}) e^{-j2\pi k x_i}\right| +\\
        &\quad \left|\sum_{i=1}^N \left( g^{k_0}\left(x_i\right) (x_i - x_{i-1}) e^{ -j2\pi k x_i } - \int_{x_{i-1}}^{x_i} g^{k_0}(t) e^{-{j2\pi kt}} dt \right) \right|.
    \end{align*}
    The proof is then completed using $(|x|+|y|)^2 \leq 2|x|^2 + 2|y|^2$.
\end{proof}

\begin{lemma}
    \label{lem:gap_order_stats}
    For any fixed gap $G > 0$, we have
    \begin{align*}
        \forall i \ \sup_{j: |j-i| \leq G} \left| x_j - x_i \right| \leq \frac{G}{N\zeta}.
    \end{align*}
\end{lemma}
\begin{proof}
We know that $x_{i+G} = F^{-1}((i+G)/N), x_{i-G} = F^{-1}((i-G)/N), x_i = F^{-1}(i/N)$. Then, we have the following:
    \begin{align*}
        \forall j > i :\  |x_j - x_i| &< |x_{j+1} - x_i|\\
        \therefore \sup_{j: |j-i| \leq G} \left| x_j - x_i \right| &= \max \left( |x_{i+G} - x_i|, |x_{i-G} - x_i| \right)
    \end{align*}
Further, $G/N = F(x_{i+G}) - F(x_i) = \int_{x_i}^{x_{i+G}} f(x) dx >  \zeta (x_{i+G} - x_i)$. The last inequality uses the definition of $x_i$ and the fact that $f(.) \geq \zeta$. 
\end{proof}

\begin{lemma}
    \label{lem:quasi_breakdown2}
    \begin{equation}
        \begin{split}
            &\left| \frac{1}{N} \sum_{i=1}^N \left( g^{k_0}\left(\frac{i}{N}\right) e^{ -\frac{j2\pi ki}{N} } - N\int_{\frac{i-1}{N}}^{\frac{i}{N}} g^{k_0}(t) e^{-{j2\pi kt}} dt \right) \right|^2 = \bigob{\frac{k_0^2}{N^2 \zeta^4}}.
        \end{split}
    \end{equation}
\end{lemma}
\begin{proof}
    \begin{align*}
        &\left|\sum_{i=1}^N \left( g^{k_0}\left(x_i\right) (x_i - x_{i-1}) e^{ -j2\pi k x_i } - \int_{x_{i-1}}^{x_i} g^{k_0}(t) e^{-{j2\pi kt}} dt \right) \right|\\
        &\mycomp{(a)}{=} \left|\sum_{i=1}^N \left( g^{k_0}\left(x_i\right) (x_i - x_{i-1}) e^{ -j2\pi k x_i } - (x_i - x_{i-1}) g^{k_0}(\hat{t}_i) e^{-{j2\pi k \hat{t}_i}} \right) \right|\\
        &\leq \sum_{i=1}^N \left| \left( g^{k_0}\left(x_i\right) (x_i - x_{i-1}) e^{ -j2\pi k x_i } - (x_i - x_{i-1}) g^{k_0}(\hat{t}_i) e^{-{j2\pi k \hat{t}_i}} \right) \right|\\
        &\mycomp{(b)}{\leq } \sum_{i=1}^N  (x_i - x_{i-1}) \sup_{i} \left| x_i - \hat{t}_i \right| \cdot \left\|\frac{d}{dt} \Big( g^{k_0}(t) e^{-j2\pi kt} \Big) \right\|_\infty \\
        &\mycomp{(c)}{\leq }\frac{1}{N \zeta} \cdot \left\| \frac{d}{dt} \Big( g^{k_0}(t) e^{-j2\pi kt} \Big) \right\|_\infty \sum_{i=1}^N  (x_i - x_{i-1}) \\
        &= \frac{1}{N \zeta^2} \cdot \left\| e^{-j2\pi kt} \left( \frac{d}{dt}g^{k_0}(t) - j2\pi k t g^{k_0}(t) \right)\right\|_\infty \hfill \left( \because \sum_{i=1}^N  (x_i - x_{i-1}) = \frac{1}{\zeta} \textrm{ from Lemma~\ref{lem:gap_order_stats}} \right)\\
        &= \frac{1}{N \zeta^2} \cdot \left\| \sqrt{\left( \left(\frac{d}{dt}g^{k_0}(t)\right)^2 + 4\pi^2 k^2 t^2 |g^{k_0}(t)|^2
        \right)} \right\|_\infty\\
        &\mycomp{(e)}{\leq} \frac{1}{N \zeta^2} \cdot \left\| \sqrt{\left( \left(2\pi k_0\right)^2 + 4\pi^2 k_0^2 |g^{k_0}(t)|^2 \right)} \right\|_\infty\\
        &\leq \frac{2\pi k_0}{N \zeta^2} \left(1 + \left\|g^{k_0}(t)\right\|_\infty \right) \mycomp{(f)}{\leq} \frac{2\pi k_0}{N \zeta^2} \left(1 + 1 +  \bigo \left(\frac{1}{N}\right)\right)\\
        &= \bigob{\frac{k_0}{N \zeta^2}} + \bigob{\frac {k_0}{N^2 \zeta^2}} = \bigob{\frac{k_0}{N \zeta^2}},
    \end{align*}
where (a) and (b) follow by using Lagrange's mean value theorem with $\hat{t}_i \in [x_{i-1}, x_i]$, (c) follows because $|x_i-\hat{t}_i| \leq |x_i - x_{i-1}| \leq \dfrac{1}{N \zeta}$ from Lemma~\ref{lem:gap_order_stats}, and (e) follows from $|k| \leq k_0, 0 \leq t \leq 1$ and Eqn.~(\ref{eqn_g_k0_smooth}). The inequality (f) follows from Eqn.~(\ref{eqn_g_gk0_close_at_t}) in Lemma~\ref{lem:g_gk0_close} and $|g(t)|\leq 1$.
\end{proof}

\begin{lemma}
    \label{lem:quasi_breakdown1}
    \begin{equation}
        \xpcttnb{\left| \frac{1}{N} \sum_{i=1}^N \left(g(t_i)-g^{k_0}\left(\frac{i}{N}\right)\right) e^{-\frac{j2\pi ki}{N}}\right|^2} = \bigo \left(\dfrac{1}{N^2\zeta^2} + \dfrac{k^2_0}{N \zeta^4}\right).
    \end{equation}
\end{lemma}

\begin{proof}
    \begin{align*}
        & \left| \sum_{i=1}^N \left(g(t_i)-g^{k_0}\left(x_i\right)\right) (x_i - x_{i-1}) e^{-j2\pi k x_i}\right|\\
        &\leq \sum_{i=1}^N \left|\left(g(t_i)-g^{k_0}\left(x_i\right)\right) (x_i - x_{i-1}) e^{-j2\pi k x_i}\right|\\
        &= \sum_{i=1}^N \left|\left(g(t_i)-g^{k_0}\left(x_i\right)\right) (x_i - x_{i-1})\right| \\
        &\leq \sum_{i=1}^N \left(\left|\left(g(t_i) - g^{k_0}(t_i) \right)  (x_i - x_{i-1}) \right| + \left|\left(g^{k_0}(t_i)-g^{k_0}\left(x_i\right)\right)  (x_i - x_{i-1}) \right| \right)\\
        &\mycomp{(a)}{\leq} \sum_{i=1}^N \bigo \left( \frac{1}{N} \right)  (x_i - x_{i-1}) + \sum_{i=1}^N \left|\left(g^{k_0}(t_i)-g^{k_0}\left(x_i\right)\right)  (x_i - x_{i-1}) \right|\\
        &\mycomp{(b)}{=} \bigo \left( \frac{1}{N\zeta} \right) + \ \sum_{i=1}^N \left|\left(g^{k_0}(t_i)-g^{k_0}\left(x_i\right)\right)  (x_i - x_{i-1}) \right|
    \end{align*}
    where (a) follows from Eqn.~\ref{eqn_g_gk0_close_at_t} and (b) follows because $\sum_{i=1}^N (x_i - x_{i-1}) = \dfrac{1}{\zeta}$ from Lemma~\ref{lem:gap_order_stats}.

    \begin{align*}
        &\therefore \xpcttnb{ \left| \sum_{i=1}^N \left(g(t_i)-g^{k_0}\left(x_i\right)\right) (x_i - x_{i-1}) e^{-j2\pi k x_i}\right|^2}\\
        &\mycomp{(c)}{=} \bigo \left( \frac{1}{N^2 \zeta^2} \right) + 2 \xpcttnb{\left(\sum_{i=1}^N \left|\left(g^{k_0}(t_i)-g^{k_0}\left(x_i\right)\right)  (x_i - x_{i-1}) \right|\right)^2}\\
        &=\bigo \left( \frac{1}{N^2 \zeta^2} \right) + 2 N \sum_{i=1}^N  (x_i - x_{i-1})^2 \xpcttnb{\left( \left|\left(g^{k_0}(t_i)-g^{k_0}\left(x_i\right)\right) \right|\right)^2}\\
        &\mycomp{(d)}{=} \bigo \left( \frac{1}{N^2 \zeta^2} \right) + 2N \bigo \left( \dfrac{k^2_0}{\zeta^2 N} \right) \sum_{i=1}^N (x_i - x_{i-1})^2\\
        &= \bigo \left( \frac{1}{N^2 \zeta^2} + N\dfrac{k^2_0}{\zeta^2 N}\dfrac{N}{N^2\zeta^2}\right)\\
        &= \bigo \left(\dfrac{1}{N^2\zeta^2} + \dfrac{k^2_0}{N \zeta^4}\right),
    \end{align*}
    where (c) follows because $(x+y)^2 \leq 2x^2 + 2y^2$ and because $(\sum_{i=1}^N x_i)^2 \leq N \sum_{i=1}^N x_i^2$, and (d) uses Eqn.~(\ref{eqn_g_k0_difference}).
\end{proof}

\begin{lemma}
    \label{lem:second_term_bound}
    \begin{equation}
        \xpcttnb{\|g^{k_0}(t) - \hat{g}^{k_0}(t)\|_2^2} = \bigob{\frac{k_0^3}{N\zeta^4} + \frac{k_0}{N^2\zeta^2}}
    \end{equation}
\end{lemma}
\begin{proof}
    Taking expectation in Lemma~\ref{lem:quasi_breakdown}, and the using Lemmas~\ref{lem:quasi_breakdown2}, \ref{lem:quasi_breakdown1}, we get:
    \begin{align*}
        \xpcttnb{\left| \hat{a}_k - a_k \right|^2} &\leq \bigo \left( \frac{k_0^2}{N}\right) + \bigo \left( \frac{k_0^2}{N^2}\right) = \bigo \left( \frac{k_0^2}{N \zeta^4} + \frac{1}{N^2\zeta^2} + \frac{k^2_0}{N^2\zeta^4}\right) = \bigo \left( \frac{k_0^2}{N \zeta^4} + \frac{1}{N^2\zeta^2}\right).
    \end{align*}
    Applying Lemma~\ref{lem:parseval_breakdown}, i.e. multiplying both sides by $k_0$, then completes the proof.
\end{proof}

Combining Lemmas~\ref{lem:overall_breakdown}, \ref{lem:g_gk0_close}, \ref{lem:noise_term_bound}, \ref{lem:third_term_bound}, \ref{lem:second_term_bound} completes the proof for Thm.~\ref{thm:1d_reconstruction}. That is, we obtain the following final result:
\begin{align*}
    &\xpcttnb{\|g(t) - \Tilde{g}^{k_0}(t)\|_2^2} = 
    \bigob{\frac{1}{N^2}} + \bigob{\frac{\sigma^2 k_0}{N\zeta^2}} + \bigob{\frac{N_{\bar{\delta}}k_0}{N \zeta^2}} + \bigob{\frac{k_0}{N^2 \zeta^2}} + \bigob{\frac{k^3_0}{N \zeta^4} + \frac{k^3_0 \bar{\delta}^2}{N^2 \zeta^4}} \\
    &\qquad+  \bigob{\frac{k^3_0}{N \zeta^4}} + \bigob{\frac{k_0}{N^2 \zeta^2}} \\
    &\qquad= \bigob{\frac{\sigma^2 k_0}{N\zeta^2}} + \bigob{\frac{N_{\bar{\delta}}k_0}{N \zeta^2}} + \bigob{\frac{k_0}{N^2 \zeta^2}} + \bigob{\frac{k^3_0}{N \zeta^4} + \frac{k^3_0 \bar{\delta}^2}{N^2 \zeta^4}}.
\end{align*}

\section{Proofs of Results for 2D Image Reconstruction}
\subsection{Proof of Lemma~\ref{lem:fr_ring_is_qbl}}
\begin{proof}
    Eqn. (21a), Sec. 3 in \cite{Pan1998} states that for all $k > \frac{2\pi \nu_1 r_0}{\gamma}$, the $k$-th Fourier series coefficient of $\mathcal{F}R_g(\nu_1, \cdot)$ is less than or equal to $\frac{\pi a_m r_0^2}{(1-\gamma^2)^{1/4} \sqrt{2\pi k}}e^{-\gamma k}$ for a constant $\gamma = 0.765$. When $k > \frac{2\pi \nu_1 r_0}{\gamma}$, we have $\frac{\pi a_m r_0^2}{(1-\gamma^2)^{1/4} \sqrt{2\pi k}}e^{-\gamma k} \leq de^{-\gamma k}$ for $d=\frac{\pi a_m r_0^2}{(1-\gamma^2)^{1/4} \sqrt{2\pi \frac{2\pi \nu_1 r_0}{\gamma}}}$. Note that $d$ depends on $\gamma$ and has to be independent of only frequency $k$. Eqn. 7 in \cite{Pan1998} also shows the symmetry of the Fourier series coefficients of $\mathcal{F}R_g(\nu_1,\cdot)$. A QBL signal needs to have decaying coefficients in both positive and negative $k$ axis. Hence, this symmetry can be used to argue that the decay occurs when $|k|$ increases (i.e., on the positive as well as negative $k$-axis). Hence, $\mathcal{F}R_g(\nu_1, \cdot)$ is a QBL signal. This result is proved in \cite{Pan1998} by expressing the $k$-th Fourier series coefficient of $\mathcal{F}R^{\theta}_g(\nu_1)$ (in fixed $\nu_1$) as an integral over the 2D signal $g$, multiplied with the $k$-th order Bessel function of the first kind.
\end{proof}

\subsection{Proof of Lemma~\ref{lem:fr_ring_reconstruction}}
\begin{proof}
    For each (noisy) projection in $P^{\prime}$, we can calculate its 1D Fourier transform yielding $\mathcal{F}R^{\theta_{h(1)}}_g(\cdot)+\epsilon_1(\cdot), \mathcal{F}R^{\theta_{h(2)}}_g(\cdot)+\epsilon_2(\cdot), \cdots, \mathcal{F}R^{\theta_{h(N)}}_g(\cdot)+\epsilon_N(\cdot)$\footnote{Note that the Fourier transforms of the additive noise components in the projection vectors at different angles are also statistically independent of each other.}. Consider $g_{\nu_1}(\theta) = \mathcal{F}R^{\theta}_g(\nu_1)$ (equivalently, a ring in $\mathcal{F}T_g$ with the same radius) such that $\mathcal{F}R^{\theta_{h(1)}}_g(\nu_1)+\epsilon_1(\nu_1), \mathcal{F}R^{\theta_{h(2)}}_g(\nu_1)+\epsilon_2(\nu_1), \cdots, \mathcal{F}R^{\theta_{h(N)}}_g(\nu_1)+\epsilon_N(\nu_1)$ are noisy samples of $g_{\nu_1}$. Then $g_{\nu_1}$ is a periodic function with period $2\pi$. Note that the properties of $g$ imply $\|g\|_1 = \int \int_{\Omega} |g(x, y)|dy dx \leq a_m \cdot \pi r_0^2$. As $\|\mathcal{F}T_g\|_\infty \leq \|g\|_1$, $\mathcal{F}T_g$ and $g_{\nu_1}$ are bounded. Lemma~\ref{lem:fr_ring_is_qbl} further implies that $g_{\nu_1} \sim \textrm{qbl}(k_1=\frac{2\pi \nu_1 r_0}{0.765}, \gamma=0.765)$. Applying Thm.~\ref{thm:1d_reconstruction} to bounded, periodic, quasi-bandlimited signal $g_{\nu_1}$ completes the proof, where the role of the unknown sample locations $\{t_i\}_{i=1}^N$ is played by the unknown angles $\{\theta_i\}_{i=1}^N$. Note that the period is $2\pi$, and hence the reconstruction formulae have been adjusted accordingly. It is simple to check that the different period and bound on $g_{\nu_1}$ change the expected reconstruction error by only a constant multiplicative factor.
\end{proof}

\subsection{Proof of Corollary~\ref{cor:disc_recon}}
\begin{proof}
This proof follows trivially from Lemma~\ref{lem:fr_ring_reconstruction} followed by a definite integral between 0 and $\nu_0$. Note that $N > e^{2\pi \nu_0 r_0} \implies N > e^{2\pi \nu r_0} \ \forall \nu \in (0, \nu_0)$.
\end{proof}

\subsection{Proof of Lemma~\ref{lem:sobolev}}
\begin{proof}
    This result is stated and proved in \cite[Theorem 1]{Beckmann2015}. 
    Define $\mathbb{U} := \{(u,v)\|u^2+v^2 > \nu^2_0\}$. Essentially, we have:
    \begin{align}
        e_g(\nu_0) &:= \frac{1}{2\pi} \int_{\mathbb{U}} |\mathcal{F}T_g(u, v)|^2 dv du\\
        &= \frac{1}{2\pi} \int_{\mathbb{U}} (1 + u^2 + v^2)^\alpha (1 + u^2 + v^2)^{-\alpha} |\mathcal{F}T_g(u, v)|^2 dv du\\
        &\leq \frac{1}{2\pi} \int_{\mathbb{U}} (1 + u^2 + v^2)^\alpha (1 + \nu_0^2)^{-\alpha} |\mathcal{F}T_g(u, v)|^2 dv du\\
        &\leq (1 + \nu_0^2)^{-\alpha} \|g\|_\alpha^2 \leq (\nu_0)^{-2\alpha} \|g\|_\alpha^2.
    \end{align}
The last equality follows from the definition of $\|g\|_\alpha^2$.
\end{proof}
\subsection{Proof of Theorem~\ref{thm:2d_reconstruction}}
\begin{proof}
Consider the definition of $e_g(\nu_0)$ in Lemma~\ref{lem:sobolev}. Then, we have:
    \begin{align*}
        &\xpcttnb{\| g - \hat{g} \|^2} = \xpcttnb{\| \mathcal{F}T_g - \widehat{\mathcal{F}T}_g \|^2}\\
        &= \xpcttnb{\int_0^\infty \int_0^{2\pi} (\mathcal{F}T_g(\nu \cos\theta, \nu \sin\theta) - \widehat{\mathcal{F}T}_g(\nu \cos\theta, \nu \sin\theta))^2 \nu d\theta d\nu}\\
        &\mycomp{(a)}{=} \xpcttnb{ \int_0^{\nu_0} \int_0^{2\pi} (\mathcal{F}T_g(\nu \cos\theta, \nu \sin\theta) - \widehat{\mathcal{F}T}_g(\nu \cos\theta, \nu \sin\theta))^2 \nu d\theta d\nu}\\
        &\qquad + \int_{\nu_0}^\infty \int_0^{2\pi} (\mathcal{F}T_g(\nu \cos\theta, \nu \sin\theta))^2 \nu d\theta d\nu\\
        &\leq \nu_0 \xpcttnb{\int_0^{\nu_0} \int_0^{2\pi} (\mathcal{F}T_g(\nu \cos\theta, \nu \sin\theta) - \widehat{\mathcal{F}T}_g(\nu \cos\theta, \nu \sin\theta))^2 d\theta d\nu}\\
        &\qquad + 2\pi e_g(\nu_0)\\
        &\mycomp{(b)}{=} \nu_0 \int_0^{\nu_0} \xpcttnb{\int_0^{2\pi} (\mathcal{F}T_g(\nu \cos\theta, \nu \sin\theta) - \widehat{\mathcal{F}T}_g(\nu \cos\theta, \nu \sin\theta))^2 d\theta} d\nu\\
        &\qquad + 2\pi e_g(\nu_0)\\
        &= \nu_0 \int_0^{\nu_0} \xpcttnb{\left\| \mathcal{F}R_g^{\theta}(\nu) - \widehat{\mathcal{F}R}^{\theta}_g(\nu) \right\|_2^2} d\nu+ 2\pi e_g(\nu_0)\\
&\mycomp{(c)}{=}\bigob{\frac{\nu^2_0(\sigma^2 + N_{\bar{\delta}})\log N}{N \zeta^2} + \frac{\nu_0 \log^3 N}{N \zeta^4} + \frac{\nu_0 \log^3 N \bar{\delta}^2}{N \zeta^4}} + \bigob{\nu_0^{-2\alpha}\|g\|_\alpha^2}.        
    \end{align*}
    where (a) follows by the definition of $\widehat{\mathcal{F}T}_g$, (b) follows from Fubini's theorem (since $\mathcal{F}T_g$, $\widehat{\mathcal{F}T}_g$ are bounded), and (c) follows from Corollary~\ref{cor:disc_recon} and Lemma~\ref{lem:sobolev}.
\end{proof}

\section{A Note on the Laplacian Eigenmaps Algorithm}
To make the paper as self-contained as possible, we here briefly describe the Laplacian Eigenmaps technique and its application in the context of 2D UVT. For a more detailed exposition of this technique, we refer the reader to \cite{Belkin2001} or \cite{Coifman2008} for more details.

Laplacian Eigenmaps (LE) is a non-linear dimensionality reduction technique proposed in \cite{Belkin2001}. It was used for the 2D UVT problem in \cite{Coifman2008}. Let the projection vectors of a $d \times d$ image $g$ be denoted by $\boldsymbol{w_1}, \boldsymbol{w_2}, ..., \boldsymbol{w_N}$ respectively where each $\boldsymbol{w_i} \in \mathbb{R}^d$. Note that the corresponding angles of projection are unknown. In the LE technique, a $N \times N$ graph adjacency matrix $\boldsymbol{W}$ is created such that $W_{ij} = \exp{\left(-\|\boldsymbol{w_i}-\boldsymbol{w_j}\|^2_2/\beta\right)}$ where $\beta$ is a smoothness parameter. Note that higher/lower values of $W_{ij}$ indicate greater/lower similarity between the projections $\boldsymbol{w_i}$ and $\boldsymbol{w_j}$. Given the matrix $\boldsymbol{W}$, the graph Laplacian matrix $\boldsymbol{L}$ can be computed in the form $\boldsymbol{L} = \boldsymbol{D} - \boldsymbol{W}$ where $D_{ii} = \sum_{j=1, j \neq i}^N W_{ij}$. 

LE is a neighborhood-preserving dimensionality reduction technique. Let $\boldsymbol{y_1}, \boldsymbol{y_2}, ..., \boldsymbol{y_N}$ respectively be the lower-dimensional embeddings of the projection vectors $\boldsymbol{w_1}, \boldsymbol{w_2}, ..., \boldsymbol{w_N}$. Note that each $\boldsymbol{y_i} \in \mathbb{R}^{\tilde{d}}$ where $\tilde{d} < d$. LE seeks to find $\{\boldsymbol{y_i}\}_{i=1}^N$ such that the embeddings $\boldsymbol{y_i}, \boldsymbol{y_j}$ are similar for similar projection vectors $\boldsymbol{w_i}, \boldsymbol{w_j}$. This is achieved by minimizing the following cost function
\begin{equation}
J(\{\boldsymbol{y_i}\}_{i=1}^N) := \sum_{i=1}^N \sum_{j=1, j \neq i}^N W_{ij} \|\boldsymbol{y_i}-\boldsymbol{y_j}\|^2_2,
\end{equation}
subject to the constraint that $\forall i, \boldsymbol{y_i}^t \boldsymbol{D y_i} = 1$ to prevent trivial solutions. The solution to the aforementioned optimization problem reduces to a generalized eigenvalue problem. In particular, it reduces to determining the $d'$ eigenvectors of $\boldsymbol{D}^{-1} \boldsymbol{L}$ corresponding to the $d'$ smallest (non-zero) eigenvalues. In the specific case considered in our work, we have $d' = 2$. That is, we have $\forall i, \boldsymbol{y_i} := (y_{1i},y_{2i})$. Using this, we compute temporary projection angles $\vartheta_i := \arctan \frac{y_{1i}}{y_{2i}}$, and use them to sort the projections angularly along a unit circle as followed in \cite{Coifman2008}.

\begin{algorithm*}[h]
\caption{Reconstruction Algorithm (\textsf{RA})}
\label{alg:reconstruction}

\begin{algorithmic}[1]
\REQUIRE Set of $N$ noisy projections $\{\boldsymbol{q_i}\}_{i=1}^N$ at angles $\{\theta_i\}_{i=1}^N$ drawn from distribution $F(.)$
\ENSURE Reconstructed image

\STATE \textbf{Denoise} the projections $\{\boldsymbol{q_i}\}_{i=1}^N$ using a PCA-based technique \cite{Singer2013} to yield denoised projections $\{\boldsymbol{q^{dn}_i}\}_{i=1}^N$
\STATE \textbf{Perform dimensionality reduction} on the denoised projections $\{\boldsymbol{q^{dn}_i}\}_{i=1}^N$ using Laplacian eigenmaps (LE) technique \cite{Belkin2001} to yield 2D embeddings $\{\boldsymbol{y_i}\}_{i=1}^N$ where $\forall i, \boldsymbol{y_i} = (y_{1i},y_{2i})$.
\STATE \textbf{Circularly order} the lower-dimensional embeddings $\{\boldsymbol{y_i}\}_{i=1}^N$ using the angles $\{\vartheta\}_{i=1}^N$ where $\vartheta_i = \tan^{-1}(y_{2i}/y_{1i})$. 
\STATE \textbf{Assign} angle values to the circularly projections based on order statistics. That is, we assign $\forall i, \hat{\theta}_i := 2\pi F^{-1}(i/N)$, i.e. the angle estimate for $\boldsymbol{q^{dn}_i}$.
\STATE \textbf{Divide} the ordered projections into a reconstruction set $S_R$ (containing say 90\% of the ordered projections) and a validation set $S_V$ containing the remaining 10\% of the projections (disjoint from $S_R$). The aim is to reconstruct the image  using projections in set $S_R$ and using the validation set $S_V$ to select the val $\nu_0$ based on minimization of the validation error.
\STATE \textbf{Compute} 1D Fourier Transform of each projection in set $S_R$
\FOR{each radius value $\nu \leq \nu_0$}
    \STATE \textbf{Reconstruct} coefficients $a_k$ using Lemma~\ref{lem:fr_ring_reconstruction} using only those projections from $S_R$. 
    \STATE \textbf{Choose} discretization $M$ for dense coverage of angles in Fourier space
    \STATE \textbf{Reconstruct} Fourier ring values using $a_k$ and the expression for $\widehat{\mathcal{F}R}^{\theta}_g(\nu_1)$ in Lemma~\ref{lem:fr_ring_reconstruction}, setting $k_0 = k_1$ as per Lemma~\ref{lem:fr_ring_is_qbl}.
    \STATE \textbf{Perform} inverse polar Fourier transform using pyNUFFT to produce a reconstructed image $\hat{g}_{\nu}$.
    \STATE \textbf{Compute} validation error $VE_{\nu} = \sum_{i \in S_V} \|\boldsymbol{q^{dn}_i} - \mathcal{R}_{\theta_i}\hat{g}_{\nu}\|^2_2$.
\ENDFOR
\STATE \textbf{Output} the final reconstructed image to be the reconstruction $\hat{g}_{\nu}$ associated with the least validation error $VE_{\nu}$. 

\end{algorithmic}
\end{algorithm*}

\section{Additional Experimental Results}
\begin{figure*}
    \centering    
    \includegraphics[width=0.48\textwidth]{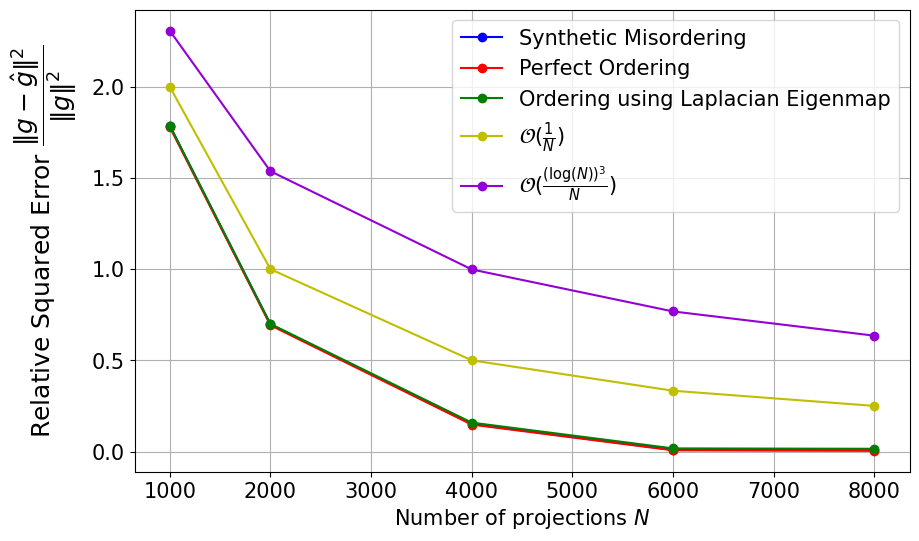}
    \includegraphics[width=0.48\textwidth]{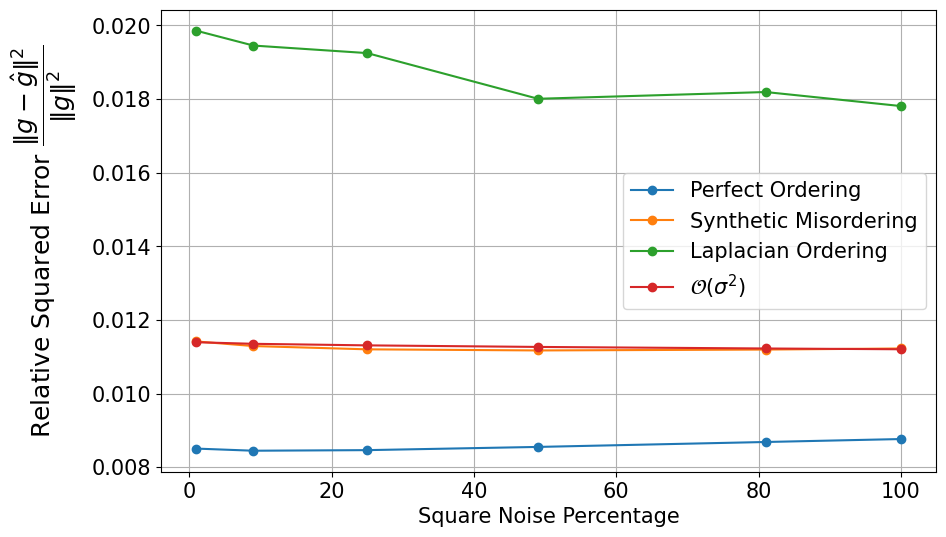}   \includegraphics[width=0.48\textwidth]{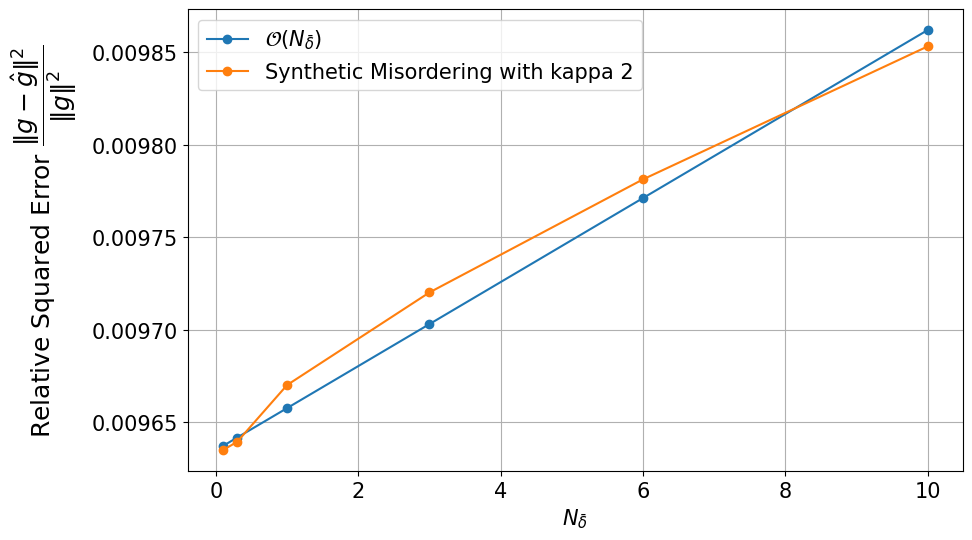}
    \caption{For an angle distribution given by $\textrm{VM}(\pi,\kappa)$ for $\kappa = 2$: Variation of $E$ versus $N$ for $f_n = 0.05$ (top left) for settings S1, S2, S3 defined in Section 4 of the main paper; Variation of $E$ versus squared noise percentage $f_n ^ 2$ for $N = 5000$ (top right) using the Laplacian eigenmaps algorithm for projection ordering; Variation of $E$ versus $N_{\bar{\delta}}$ for $N = 5000, f_n  = 0.05$ (bottom row) using the Laplacian eigenmaps algorithm for projection ordering.}
    \label{fig:kappa2}
\end{figure*}

\begin{figure*}
    \centering
    \includegraphics[width=0.125\textwidth]{paper_images/recons/ground_truth.png}
    \includegraphics[width=0.48\textwidth]{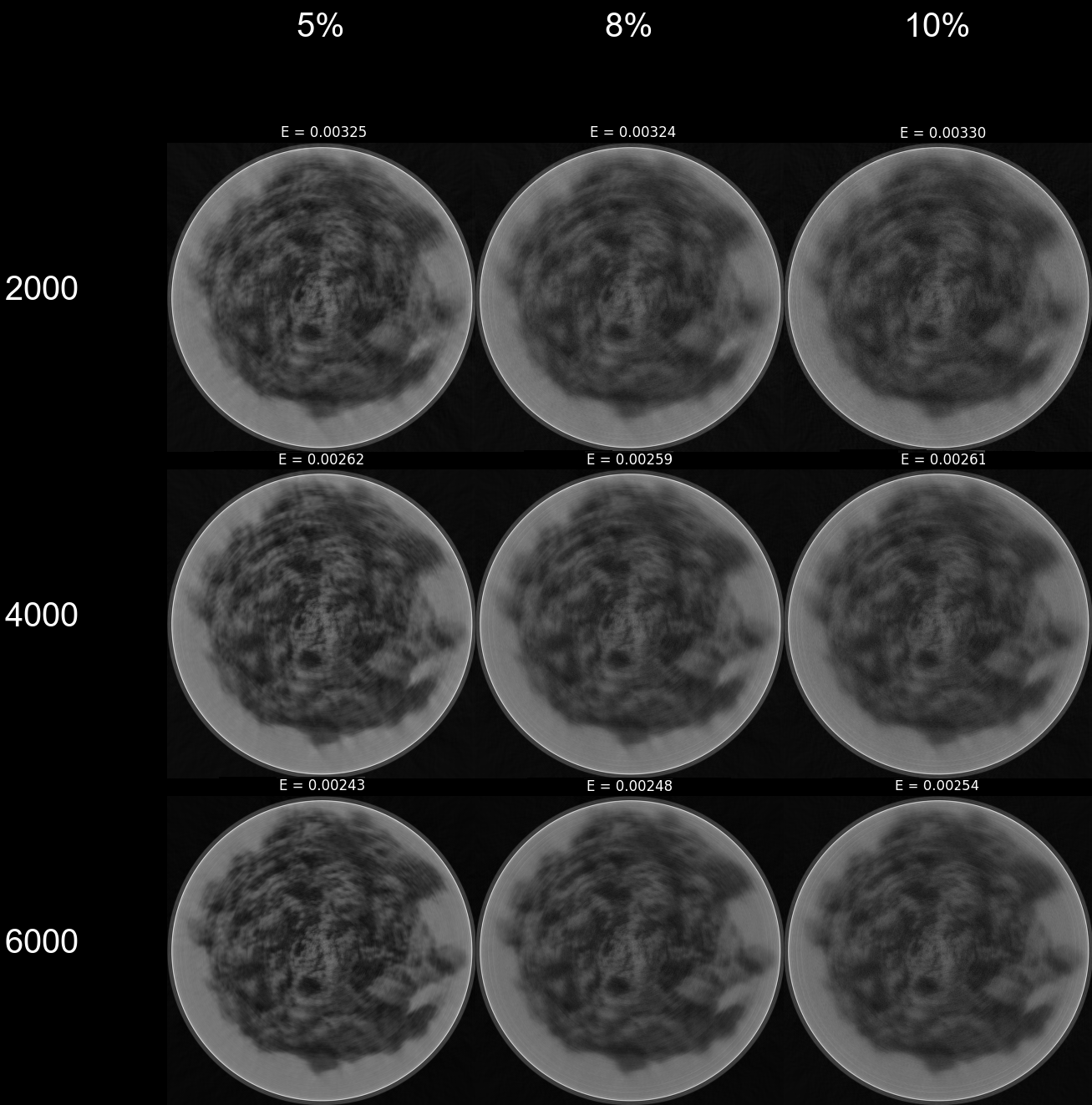}
    \includegraphics[width=0.48\textwidth]{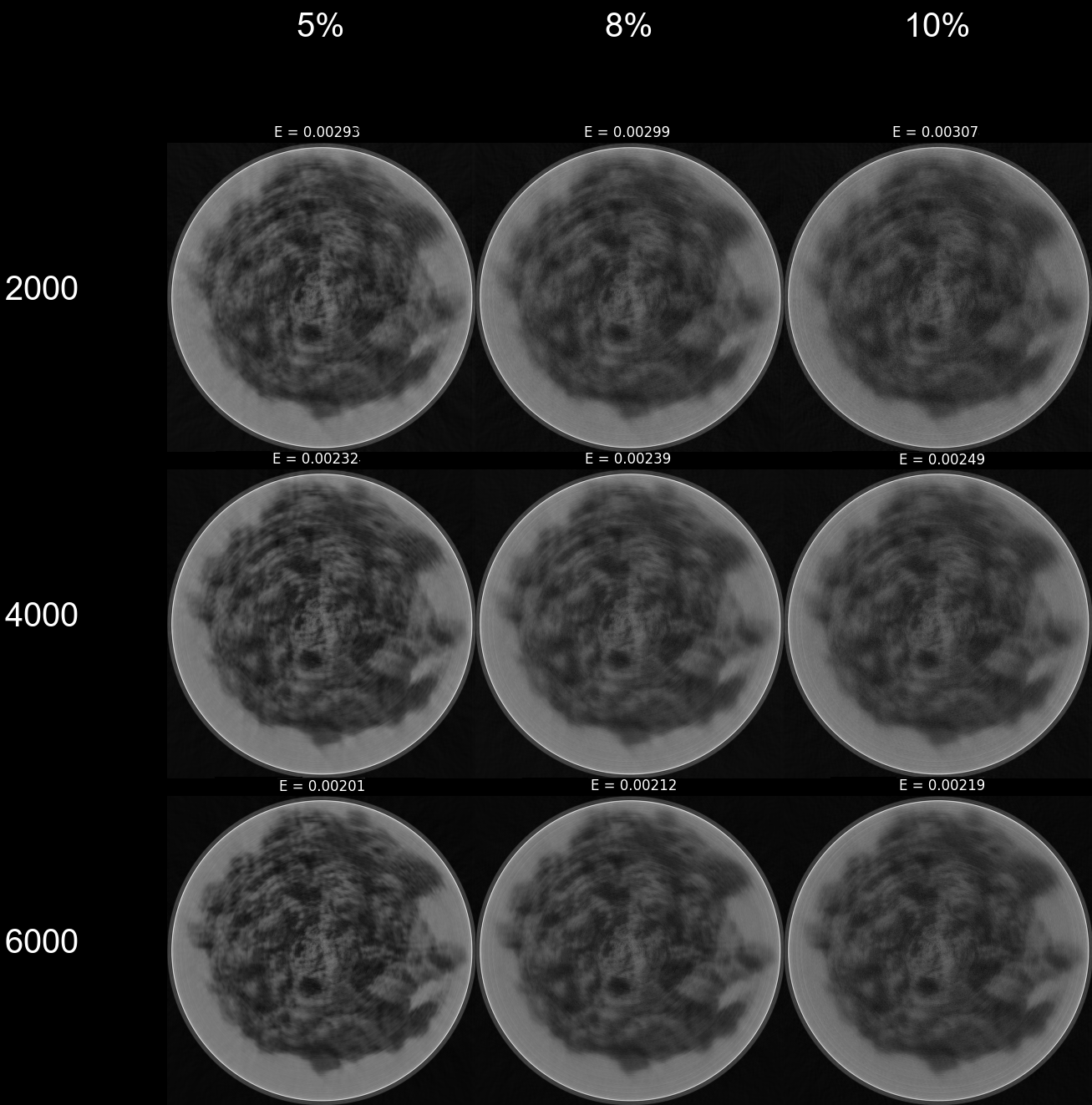}
    \includegraphics[width=0.48\textwidth]{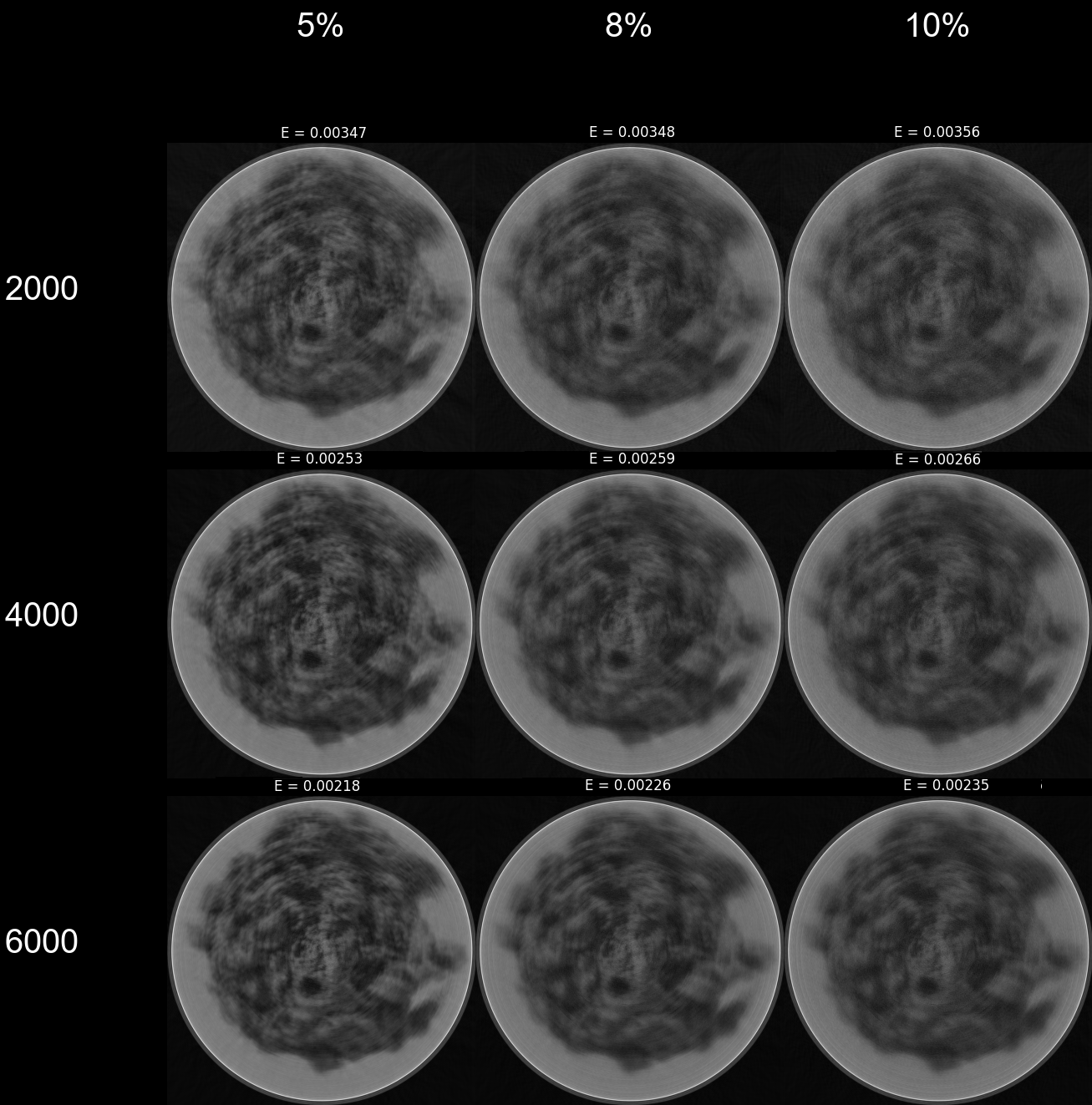}
    \caption{Leftmost image (top row): ground truth image; Sample reconstructions and relative reconstruction error ($E$) values with $N \in \{2000,4000,6000\}$ projections and $f_n \in \{0.05,0.08,0.1\}$ for setting S3 defined in Sec. 4 of the main paper, i.e. via Laplacian eigenmaps, for angles drawn from $\textrm{VM}(\pi,\kappa)$ for $\kappa = 0.33$ (top right), $\kappa = 0.5$ (bottom left) and $\kappa = 0.75$ (bottom right).}
    \label{fig:recons_other_kappa}
\end{figure*}

Additional experimental results, supplementing those in Section 4 of the main paper, are provided in this section. Refer to Fig.~\ref{fig:kappa2} for a plot of the reconstruction error versus the number of projections $N$ (keeping noise std. dev. $\sigma$ fixed at $5\%$), versus squared noise percentage $f^2_n$ (keeping $N = 5000$) and versus the number of ordering errors $N_{\bar{\delta}}$ (keeping $N = 5000, f_n = 0.05$). All these three plots are for the case when the angles of projection were drawn from $\textrm{VM}(\pi,\kappa)$ with $\kappa = 2$. Refer to Fig.~\ref{fig:recons_other_kappa} for reconstruction results for different values of $N$ and $\sigma$ for the case where the angles of projection were drawn from $\textrm{VM}(\pi,\kappa)$ where $\kappa \in \{0.33,0.5,0.75\}$.

\bibliographystyle{elsarticle-num} 
\bibliography{sample}

\end{document}